\documentclass[letterpaper, 10 pt, conference]{ieeeconf}  

\IEEEoverridecommandlockouts                              

\usepackage{graphics} 
\usepackage{epsfig} 
\usepackage{times} 
\usepackage{amsmath} 
\usepackage{amssymb}  

\usepackage{booktabs}       
\usepackage{amsfonts}       
\usepackage{nicefrac}       
\usepackage{microtype}      
\usepackage{xcolor}         

\usepackage{graphicx}
\usepackage{amsmath}
\usepackage{amssymb}
\usepackage{booktabs}
\usepackage{bbm}
\usepackage{multirow}
\usepackage{textcomp}

\usepackage[capitalize]{cleveref}
\crefname{section}{Sec.}{Secs.}
\Crefname{section}{Section}{Sections}
\Crefname{table}{Table}{Tables}
\crefname{table}{Tab.}{Tabs.}

\def\ie{\emph{i.e.}}

\def\wrt{w.r.t.}

\title{\LARGE \bf
MonoPlane: Exploiting Monocular Geometric Cues for \\Generalizable 3D Plane Reconstruction
}

\author{Wang Zhao$^{*1}$, Jiachen Liu$^{*2}$, Sheng Zhang$^{3}$, Yishu Li$^{1}$, Sili Chen$^{3}$, \\Sharon X Huang$^{2}$, Yong-Jin Liu$^{1}$ and Hengkai Guo$^{3}$
\thanks{*Equal Contribution.}
\thanks{\textdagger This work was supported by the Natural Science Foundation of China
(62332019).}
\thanks{$^{1}$ BNRist, MOEKey Laboratory of Pervasive Computing, Department of Computer Science and Technology, Tsinghua University, Beijing, China.}
\thanks{$^{2}$ College of Information Sciences and Technology, Pennsylvania State University, University Park, PA 16802, USA.}
\thanks{$^{3}$ Bytedance Inc., Beijing, China.}
}

\begin{document}

\maketitle
\thispagestyle{empty}
\pagestyle{empty}

\begin{abstract}
This paper presents a generalizable 3D plane detection and reconstruction framework named MonoPlane. Unlike previous robust estimator-based works (which require multiple images or RGB-D input) and learning-based works (which suffer from domain shift), MonoPlane combines the best of two worlds and establishes a plane reconstruction pipeline based on monocular geometric cues, resulting in accurate, robust and scalable 3D plane detection and reconstruction in the wild. Specifically, we first leverage large-scale pre-trained neural networks to obtain the depth and surface normals from a single image. These monocular geometric cues are then incorporated into a proximity-guided RANSAC framework to sequentially fit each plane instance. We exploit effective 3D point proximity and model such proximity via a graph within RANSAC to guide the plane fitting from noisy monocular depths, followed by image-level multi-plane joint optimization to improve the consistency among all plane instances. We further design a simple but effective pipeline to extend this single-view solution to sparse-view 3D plane reconstruction. Extensive experiments on a list of datasets demonstrate our superior zero-shot generalizability over baselines, achieving state-of-the-art plane reconstruction performance in a transferring setting. Our code is available at https://github.com/thuzhaowang/MonoPlane.
\end{abstract}
\section{INTRODUCTION}
\label{sec:intro}

Reconstructing 3D planes from images is a fundamental task in 3D vision that has wide applications in downstream tasks, such as robotics, mixture reality and interior modeling. The goal of this task is to simultaneously detect planes and estimate plane parameters from single or multiple images.

Traditional methods rely on multi-view cues~\cite{sinha2009piecewise}, depth sensor inputs~\cite{salas2014dense, raposo2013plane}, or certain heuristic scene assumptions~\cite{furukawa2009manhattan}, and utilize robust estimators like RANSAC~\cite{fischler1981random} and MRFs~\cite{boykov1998markov} optimization to fit plane proposals and label pixels. These methods achieve desirable plane reconstruction results in the presence of depth or multi-view input. However, their performance may drastically degrade when the heuristic assumptions are broken, or encountering low-texture images, motion blurs and noisy or incomplete depth. 

Learning-based methods have been proposed to reconstruct 3D planes from a single image~\cite{liu2019planercnn, qian2020learning, tan2021planetr}, multi-view images~\cite{liu2022planemvs, tan2022nope, agarwala2022planeformers} or videos~\cite{xie2022planarrecon}. Thanks to the construction of large-scale benchmark datasets~\cite{dai2017scannet, chang2017matterport3d}, these methods have showcased remarkable performance in segmenting plane instances and regressing plane parameters on particular dataset. To achieve better plane reconstruction, researchers have proposed to leverage more advanced network architectures~\cite{liu2019planercnn, shi2023planerectr}, or utilize geometric primitives such as line segments~\cite{tan2021planetr} to assist, or optimize multi-view planes and camera pose in a joint manner~\cite{jin2021planar, agarwala2022planeformers}. Despite the significant progress, such data-driven methods still require costly groundtruth labeling and more importantly, have shown poor generalization ability onto out-of-distribution data as shown in our experiments. However, transferring across different environments is a crucial capacity in real-world applications such as robot navigation or augmented reality.

Recently, several works explore the possibility of achieving generalizable depth or normal estimation from single-view images~\cite{ranftl2020towards, eftekhar2021omnidata, bhat2023zoedepth}. These works leverage rich depth-relevant data from different sources such as stereo, RGB-D video, 3D movies, and mix them together to obtain large-scale training datasets. These trained neural networks demonstrate superior zero-shot accuracy on in-the-wild data. Some recent studies have leveraged these pre-trained models to assist in different 3D vision tasks such as neural 3D reconstruction and rendering~\cite{Yu2022MonoSDF, zhu2023nicer}.
\begin{figure}[tb]
    \centering
    \scriptsize
    \setlength{\tabcolsep}{2pt}
    \begin{tabular}{ccc}

    \centering
        {\includegraphics[width=0.30\linewidth]{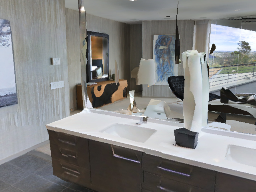}} &  
        {\includegraphics[width=0.30\linewidth]{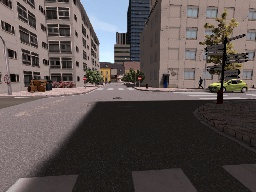}} &  
        {\includegraphics[width=0.30\linewidth]{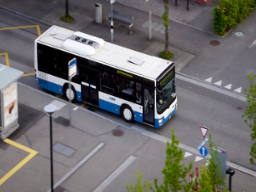}} \\
        
        {\includegraphics[width=0.30\linewidth]{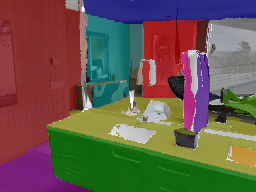}} &  
        {\includegraphics[width=0.30\linewidth]{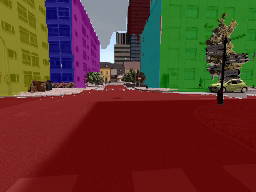}} &  
        {\includegraphics[width=0.30\linewidth]{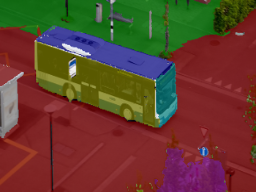}} \\

        Indoor - Matterport~\cite{chang2017matterport3d} &  Outdoor - Synthia~\cite{ros2016synthia} & In-the-wild Data~\cite{perazzi2016benchmark}

    \end{tabular}
    \caption{Sample results from diverse scenarios. Our method achieves superior generalizable 3D plane detection and reconstruction by incorporating generalizable cues and robust designs.}
    \label{fig::teaser}
    \vspace{-10pt}
\end{figure}
Inspired by these recent successes using pretrained monocular geometric models, we aim to exploit the potential of monocular geometric cues in 3D plane reconstruction.
Taking a single RGB image as input, the pretrained networks are employed to predict depth and surface normals. To make fully use of these geometric cues, we revisit the classical RANSAC~\cite{fischler1981random} pipeline, and further propose a monocular cue guided proximity modeling approach upon the graph-cut RANSAC~\cite{barath2018graph}. This specially-designed RANSAC pipeline is able to effectively utilize the monocular geometric information while being robust to handle noisy input. Then, an image-level multi-plane joint optimization is conducted via dense CRFs~\cite{krahenbuhl2011efficient} to encourage more consistent plane masks by combining the RGB color information. In this way, our method, termed MonoPlane, without plane labels and domain-specific training, effectively predicts satisfactory 3D plane reconstruction results from a single RGB image. We further demonstrate that our single-view solution can be easily extended to sparse uncalibrated images. Experimental results across different benchmarks and in-the-wild data (Figure~\ref{fig::teaser}) showcase that our proposed framework is accurate, generalizable and scalable.

To summarize, our contributions are three-fold:
\begin{itemize}
\item We explore the use of monocular geometric cues for 3D plane reconstruction, delivering generalizable 3D plane reconstruction from a single or sparse RGB images.
\item We propose a point-proximity based graph-cut RANSAC algorithm by incorporating monocular geometric cues, which greatly enhanced the robustness on handling noisy input or complex scenes.
\item We build a generalizable 3D plane reconstruction system for both single image and sparse-view images without any plane groundtruth or domain-specific finetuning. Experiments on various data from diverse environments clearly demonstrate the effectiveness of our system.
\end{itemize}
\section{RELATED WORK}
\label{sec:related_work}

\subsection{Piece-wise Plane Reconstruction from Images}


Traditional 3D plane reconstruction approaches typically use RGB-D image streams~\cite{salas2014dense, raposo2013plane} as input, segment the planes with 3D point cloud, or require multi-view images~\cite{sinha2009piecewise} to build cross-view correspondences and objectives for plane discovery. Robust estimators such as RANSAC~\cite{fischler1981random} and optimization tools such as Conditional Random Field~(CRF)~\cite{gallup2010piecewise} and Markov Random Field~(MRF)~\cite{sinha2009piecewise, furukawa2009manhattan} are commonly employed to fit the plane model and label image pixels. Although these methods achieve reasonable results on plane reconstruction, they often rely on expensive and inflexible equipment, such as multi-view setups or depth sensors, which limit their practical use in real-world applications. Besides, certain methods~\cite{furukawa2009manhattan} assume strong scene priors such as Manhattan-world stereo, thereby restricting their applicability to versatile and complex environments. In recent years, some learning-based methods have been proposed, aiming to recover 3D planes from a single RGB image in a fully supervised way. The detection and parameter estimation processes are typically performed either in a top-down~\cite{liu2018planenet, liu2019planercnn, qian2020learning}, or bottom-up~\cite{yang2018recovering, yu2019single, tan2021planetr, shi2023planerectr} manner.

\par To leverage the multi-view geometry, PlaneMVS~\cite{liu2022planemvs} proposes a slanted-hypotheses-based plane warping strategy, enabling recovering planes from multiple views with known camera poses. PlanarRecon~\cite{xie2022planarrecon} presents a voxel-based approach to incrementally track and reconstruct planes of a scene from posed monocular videos. Another line of works~\cite{jin2021planar, agarwala2022planeformers, tan2022nope} focus on reconstructing planes from sparse views with unknown camera poses. S3PRecon~\cite{ye2023self} proposes a self-supervised way to cluster 3D planes from monocular input images through neural 3D reconstruction. However, a general limitation in these learning-based methods is that they predict planar geometry using deep features trained on specific, homogeneous environments. As a result, their generalizability to novel scenes is highly constrained, especially when transferring to data with significant gaps from the training data. To resolve this bottleneck, our proposed method focuses on developing a transferable system that leverages pre-trained monocular geometric models and a robust plane estimator based on RANSAC. By adopting these techniques, we aim to avoid introducing biases associated with deep features from particular datasets and enhance the adaptability of plane reconstruction across diverse scenes.

\subsection{Generalizable Monocular 3D Estimation}
3D geometry estimation including estimating depth~\cite{eigen2014depth, eigen2015predicting, bhat2021adabins, li2023depthformer, agarwal2023attention} and surface normals~\cite{eigen2015predicting, bansal2016marr} from a single image has been studied via deep learning. These methods design effective networks such as CNNs~\cite{eigen2014depth, bansal2016marr} and Transformers~\cite{li2023depthformer, agarwal2023attention}, showing superior performance on specific benchmarks~\cite{geiger2013vision, dai2017scannet, silberman2012indoor, song2015sun}. However, their generalizability to new environments is generally limited, posing challenges when applied to real-world data. 

In recent studies, zero-shot transfer learning on 3D estimation~\cite{ranftl2020towards, eftekhar2021omnidata, ranftl2021vision, bhat2023zoedepth} is drawing increasing attention.
MiDaS~\cite{ranftl2020towards} constructs a large-scale, mixed dataset from diverse sources and introduces a scale-shift invariant training loss to learn relative depth across different annotation formats. DPT~\cite{ranftl2021vision} proposes a network which combines the advantage of vision transformers~\cite{dosovitskiy2020image} and CNNs to achieve global receptive field while preserving fine-grained details on depth estimation. Omnidata~\cite{eftekhar2021omnidata} designs a data-centric approach to generate massive high-quality training data to facilitate generalizable geometric prediction tasks. To address the limitation of previous work on merely estimating relative depth, follow-up works~\cite{bhat2023zoedepth, guizilini2023towards, yin2023metric3d} aim to learn scale-preserving metric depth through scale-adaptive strategies or further enlarged datasets across domains and datasets.

The aforementioned works serve as strong motivation for our research, inspiring us to develop a cross-dataset 3D plane reconstruction system with enhanced generalizability. Additionally, the monocular geometric cues inferred from these models, offer valuable initializations and solid foundation for our plane reconstruction pipeline. These cues provide a desirable starting point and contribute significantly to the effectiveness of our overall reconstruction process.
\section{METHOD}
\label{sec:method}
\begin{figure*}[tb]
    \centering
    \includegraphics[width=1.0\linewidth]{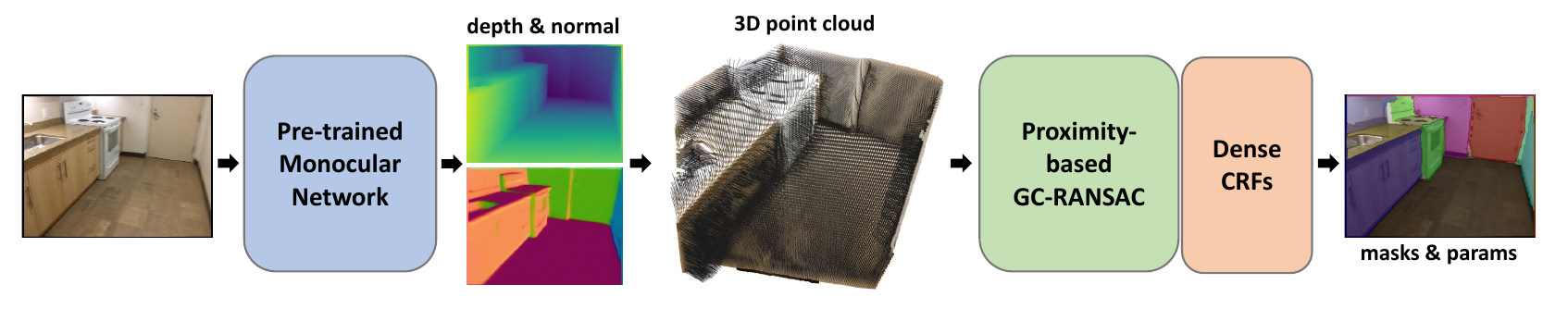}
    \caption{Pipeline of our proposed method MonoPlane. Given a single image as input, a pre-trained monocular network predicts depths and normals for each image, resulting in 3D oriented point clouds. Our proposed proximity based graph-cut RANSAC is then applied to handle noisy point clouds and sequentially segment 3D planes, followed by projection and dense CRFs to output plane masks and parameters.}
    \label{fig::pipeline_single}
    \vspace{-10pt}
\end{figure*}

\begin{figure}[tb]
    \centering
    \scriptsize
    \setlength{\tabcolsep}{2pt}
    \begin{tabular}{ccccc}

    \centering
        {\includegraphics[width=0.15\linewidth]{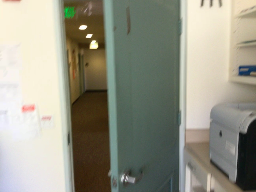}} &  
        {\includegraphics[width=0.15\linewidth]{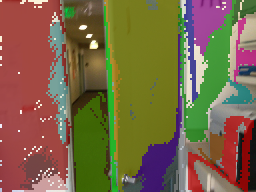}} &  
        {\includegraphics[width=0.15\linewidth]{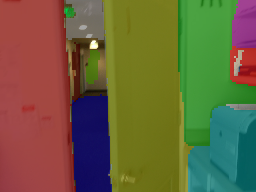}} & 
        {\includegraphics[width=0.15\linewidth]{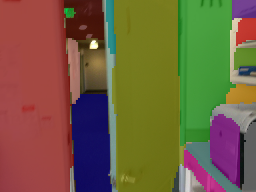}} &  
        {\includegraphics[width=0.15\linewidth]{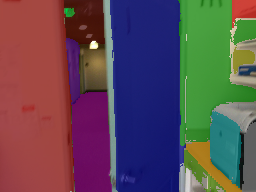}} \\  

        {\includegraphics[width=0.15\linewidth]{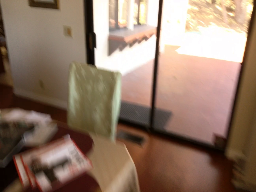}} &  
        {\includegraphics[width=0.15\linewidth]{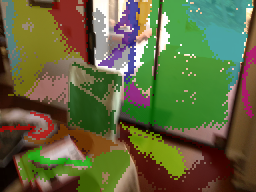}} &  
        {\includegraphics[width=0.15\linewidth]{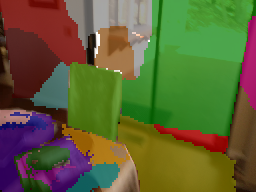}} & 
        {\includegraphics[width=0.15\linewidth]{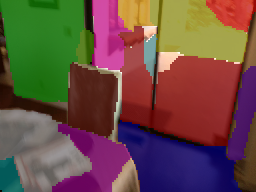}} &  
        {\includegraphics[width=0.15\linewidth]{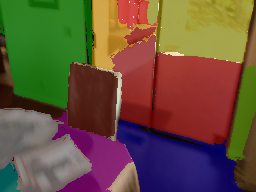}} \\

        Image & Seq-RANSAC & GC-RANSAC & Ours w/o CRF & Ours \\

    \end{tabular}
    \caption{Example results of different RANSAC baselines and our method.}
    \label{fig::ablation}
    \vspace{-10pt}
\end{figure}

Our approach endeavors to detect and reconstruct 3D planes from images in a generalizable manner. This is achieved by utilizing mix-dataset pre-trained monocular prediction models, and proximity-based RANSAC optimization. Starting with a single image with predicted monocular depth, we first demonstrate a sequential RANSAC baseline and discuss its limitations in Section~\ref{sec::baseline}. To address these limitations, we introduce a graph-based optimization method guided by monocular cues in Section~\ref{sec::graph_optim}. This method effectively capitalizes the rich proximity information of the points to determine the appropriate planes. Finally, in Section~\ref{sec::multiview}, we extend the single-image solution to sparse-view image input. An overview of our pipeline is shown in Figure~\ref{fig::pipeline_single}.

\subsection{Sequential RANSAC Baseline}
\label{sec::baseline}
Given an RGB image $I$ as input, we first utilize a large-scale pretrained model such as Omnidata~\cite{eftekhar2021omnidata} to predict the depth $D$ and surface normal $N$. Point cloud $P$ is then obtained through un-projection $P = D * K^{-1} * X$, where $X$ denotes the homogeneous pixel coordinate and $K$ is the camera intrinsic. To handle the variation of monocular depth scale, we normalize the point cloud roughly into a unit sphere. With the 3D point cloud available, a straight-forward idea is to use the RANSAC algorithm to fit 3D planes.

In the RANSAC process, the sampler first randomly selects a minimal sample of 3 points from point set $\mathcal{P} \subseteq \mathbb{R}^3$ to estimate a plane model $\theta \in \mathbb{R}^4$. Then the point-to-plane distance is used as the cost function $\phi: \mathcal{P} \times \mathbb{R}^4 \rightarrow \mathbb{R}$ to calculate all the data errors. Points with error smaller than threshold $\epsilon$ are labeled as inliers, otherwise outliers. The scorer simply assesses the inlier number to score an estimated plane model. 
This process repeats until certain iterations are reached or sufficient inliers are found. And the plane with the highest score is selected as the best fit model. This standard RANSAC procedure can also be formulated as an energy minimization problem, as stated in~\cite{barath2018graph}. Denote $L \in \{0;1\}^{|\mathcal{P}|}$ as a binary labeling (0-outlier, 1-inlier) of $|\mathcal{P}|$ points, the data energy is defined as follows:
\begin{equation}
\small{
    E_d(L) = \sum_{p\in \mathcal{P}} || L_p||_{\{0;1\}}
    }
\end{equation}
where
\begin{equation}
\small{
   ||L_p||_{\{0;1\}} = 
    \begin{cases}
		0 & \text{if } (L_p = 0 \wedge \phi(p, \theta) \geq \epsilon) \text{ } \vee \\    
		 & \text{\phantom{xx}} (L_p = 1 \wedge \phi(p, \theta) < \epsilon) \\    
        1 & \text{otherwise.}
	\end{cases}
 }
\end{equation}

Minimizing this energy leads to the same binary inlier-outlier labeling given current hypothesis model $\theta$. This standard RANSAC procedure finds a best plane model and its inlier set, one at a time. Therefore, to detect arbitrary number of plane instances, we sequentially apply the RANSAC on the remaining outlier points of current step. 

We test this sequential RANSAC for 3D plane fitting from monocular depths, as shown in Figure~\ref{fig::ablation}. Sequential RANSAC is generally able to detect the dominant planes, however, the plane detection results contain massive outliers and small pieces of plane fragments. The main reason is that, the predicted monocular depths, although depicting the rough geometric structures, still involve non-negligible noise. The noise distribution is uneven, hard to model and not tolerable with a single global inlier-outlier threshold, making this simple RANSAC baseline impractical for real-world use. To this end, we propose to utilize the point-to-point proximity in graphs, which assists to reduce the effects from depth noise.

\subsection{Graph Optimization with Monocular Cues}
\label{sec::graph_optim}
The vanilla RANSAC treats each point independently, \ie, each point is labeled as inlier or outlier solely based on its own point-to-plane distance regardless of the other points. However, a 3D plane, as a geometric primitive, is defined locally thus naturally conforms to the spatial coherence constraints. Thus, we aim to explore point relations to assist with RANSAC labeling. Inspired by this observation, in this section we solve the RANSAC energy optimization problem using graphs to utilize point proximity information.

\subsubsection{Sequential Graph-cut RANSAC for Plane Fitting}
 In RANSAC community, Graph-cut RANSAC~\cite{barath2018graph} shares this insight and models the spatial conherence with a graph. GC-RANSAC has shown impressive results on several geometric tasks such as fundamental/essential matrix estimation, homography and affine transformation estimation. We further extend it to 3D plane fitting. Specifically, GC-RANSAC introduces a neighborhood graph $\mathcal{G}=(\mathcal{P}, \mathcal{E})$, and then utilizes a new pairwise energy term $E_s$ in addition to the data energy term:
 \begin{equation}
\small{
   E_s = \sum_{p,q\in \mathcal{E}}
    \begin{cases}
		1 & \text{if } L_p \neq L_q \\    
		  0 & \text{if } L_p = L_q = 1 \\    
            \frac{1}{2} (f(0, p) + f(0, q)) & \text{if } L_p = L_q = 0 \\
	\end{cases}}
\end{equation}
where
\begin{equation}
\small{
   f(L_p, p) = 
    \begin{cases}
		0 & \text{if } (L_p = 0 \wedge \phi(p, \theta) \geq \epsilon) \text{ } \\ & \text{   }  \vee (L_p = 1 \wedge \phi(p, \theta) < \epsilon) \\ 
		  \frac{\phi^2(p, \theta)}{\epsilon^2} & \text{if } L_p = 1 \wedge \phi(p, \theta) \geq \epsilon \\    
            1 - \frac{\phi^2(p, \theta)}{\epsilon^2} & \text{if } L_p = 0 \wedge \phi(p, \theta) < \epsilon \\
	\end{cases}}
\end{equation}
is the cost function with truncated $L_2$ loss used in MSAC~\cite{torr2002bayesian}. Note that other alternative cost functions can also be adopted, such as the Gaussian-kernel function~\cite{torr2000mlesac} or MAGSAC++~\cite{barath2020magsac++} loss function. Please refer to \cite{barath2018graph} for more details. This pairwise energy $E_s$, following the Potts model~\cite{boykov1998markov}, is designed to penalize differently labeled neighbors, and avoid penalizing neighboring points $p$ and $q$ if they are both labeled as inliers, thus encouraging the spatial coherence of the inlier set.
The data term is $E_d = \sum_{p\in \mathcal{P}} f(L_p, p)$ and the final energy $E$ is defined as:
\begin{equation}
\small{
   E = (1-\lambda) E_d + \lambda E_s \\}
\end{equation}
where $\lambda$ is a hyper-parameter to control the smoothness. This optimization can be solved efficiently and globally via the graph-cut algorithm. The desired globally optimal labeling $L^* = \arg \min_{L} E(L)$ is determined in polynomial time. 

We implement the sequential GC-RANSAC and test it for 3D plane fitting from monocular depths. Figure~\ref{fig::ablation} shows that the GC-RANSAC greatly improves the spatial coherence. However, it still suffers from over-smoothness or under-smoothness given noisy monocular depths, and is sensitive to the balancing weight $\lambda$ and the distance threshold $\epsilon$.

\subsubsection{Point Proximity Modeling with Monocular Cues}
To tackle the above issues, we introduce point-to-point proximity inside GC-RANSAC to better model the point relations. The insight is simple: In GC-RANSAC, giving different labels to neighboring points is penalized via a constant energy term, regardless of the points' similarity. However, we propose that the penalty should be conditioned on the probability of neighboring points being labeled as the same. For example, if two points are neighbors spatially but have distinct colors and surface normals, labeling these two points as different is supposed to be reasonable and the penalty should be small. This proximity modeling has been widely explored in vision community such as markov random fields~\cite{krahenbuhl2011efficient}, bilateral filtering~\cite{tomasi1998bilateral}, etc. Different from previous works, our framework introduces proximity modeling into the RANSAC procedure via Graph-cut RANSAC, and enhances the modeling with additional monocular cues obtained from pretrained neural networks. 

Specifically, we formulate three proximity terms $k_p, k_c, k_n$ from measures of similarity about positions, colors, and surface normals. These terms are formulated like Gaussian kernels. For two neighboring points with 3D positions $p_i, p_j$, image color vectors $I_i, I_j$ and surface normals $n_i, n_j$, the proximity energy of each term is defined as:
\begin{align}
\label{eq::kp}
    & k_p = w_p \exp \left( -\frac{|p_i - p_j|^2}{2\alpha^2_p} \right) \\
    \label{eq::kc}
    & k_c = w_c \exp \left( -\frac{|p_i - p_j|^2}{2\alpha^2_p}  - \frac{|I_i - I_j|^2}{2\alpha^2_c}\right) \\
    \label{eq::kn}
    & k_n = w_n \exp \left( -\frac{|p_i - p_j|^2}{2\alpha^2_p}  - \frac{\arccos{(n_i \cdot n_j)}}{2\alpha^2_n}\right)
\end{align}

Now the pairwise energy term $E_s$ is formulated as:
\begin{equation}
\small{
   E_s = \sum_{p,q\in \mathcal{E}}
    \begin{cases}
		  0.5 + k_p + k_c + k_n & \text{if } L_p \neq L_q \\    
		  0 & \text{if } L_p = L_q = 1 \\    
            \frac{1}{2} (f(0, p) + f(0, q)) & \text{if } L_p = L_q = 0 \\
	\end{cases}}
\end{equation}

The graph-cut optimization solver \cite{kolmogorov2004energy} used in GC-RANSAC requires the inequality $e_{pq}(0,0) + e_{pq}(1,1) \leq e_{pq}(1,0) + e_{pq}(0,1)$ to hold. Since $k_p, k_c, k_n > 0$ and $f(0, \cdot) \leq 1$, our pairwise energy meets the sub-modular requirements.

As for the data energy term, we inherit $E_d$ of the GC-RANSAC and further utilize surface normal to define an additional data energy term. The observation is that the predicted monocular surface normals should show strong homogeneity in planar regions, thus can be used for rejecting outlier points based on normal similarity. The proposed normal data energy is $E_n = \sum_{p \in \mathcal{P}} g(L_p, p)$, where
\begin{equation}
\small{
   g(L_p, p) = 
    \begin{cases}
		  \exp \left( \frac{\rho(p, \theta_n)}{\epsilon_n} - 1 \right) & \text{if } L_p = 1 \wedge \rho(p, \theta_n) \geq \epsilon_n \\
            0 & \text{otherwise.}
	\end{cases}
 }
\end{equation}
$\rho: \mathcal{P} \times \mathbb{R}^3 \rightarrow \mathbb{R}$ denotes the normal error function, which calculates the normal angle difference between point $p$ and current model normal $\theta_n$. $\epsilon_n$ is the normal angle threshold. In this energy term, we only penalize those points that are labeled as inliers but still have large normal errors. Exponential cost prevents data points whose normals are very different from the model normal from being labeled as inliers. Note that the model normal $\theta_n$ is the average of monocular normal predictions of current minimal or non-minimal samples. The overall optimization energy is then:
\begin{equation}
\small{
    E = (1-\lambda)E_d + (1-\lambda)E_n + \lambda E_s
    }
\end{equation}

\begin{figure}[tb]
    \centering
    \includegraphics[width=1.0\linewidth]{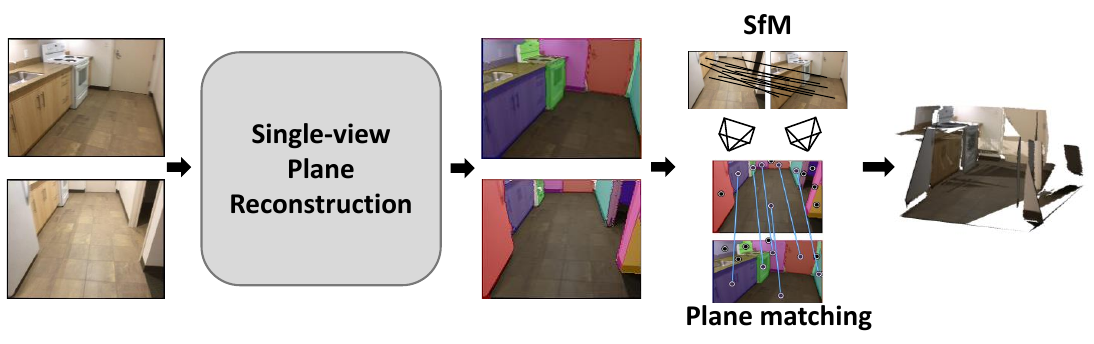}
    \caption{Extension to sparse views. We first utilize our single-view pipeline to get 3D plane proposals for each image. These plane proposals are then matched and fused into global 3D planes.}
    \label{fig::pipeline_sparse}
    \vspace{-15pt}
\end{figure}
Finally, we utilize the surface normal homogeneity to help the RANSAC sampler. Given a minimal data sample, we calculate the normal angle differences among each pair of points. If any pair of points has greater angle difference than threshold $\epsilon_n$, this minimal sample will be abandoned. This normal check can effectively reject bad minimal samples and accelerate the RANSAC procedure.

As shown in Figure~\ref{fig::ablation}, our method greatly improves over the previous baselines, producing accurate and smooth plane detections. Thanks to point proximity modeling guided by monocular cues, our method can handle imperfect inputs and find the most fit plane model with the best inlier set.

\begin{figure*}[tb]
    \centering
    \scriptsize
    \setlength{\tabcolsep}{2pt}
    \begin{tabular}{ccccccc}

    \centering
        {\includegraphics[width=0.125\linewidth]{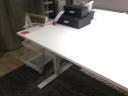}} &  
        {\includegraphics[width=0.125\linewidth]{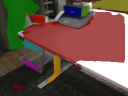}} &  
        {\includegraphics[width=0.125\linewidth]{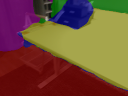}} & 
        {\includegraphics[width=0.125\linewidth]{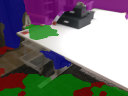}} &
        {\includegraphics[width=0.125\linewidth]{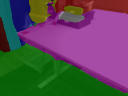}} &
        {\includegraphics[width=0.125\linewidth]{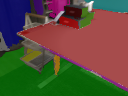}} &
        {\includegraphics[width=0.125\linewidth]{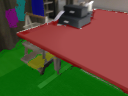}}\\

        {\includegraphics[width=0.125\linewidth]{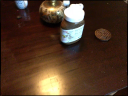}} &  
        {\includegraphics[width=0.125\linewidth]{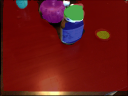}} &  
        {\includegraphics[width=0.125\linewidth]{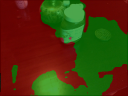}} & 
        {\includegraphics[width=0.125\linewidth]{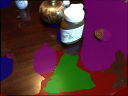}} &
        {\includegraphics[width=0.125\linewidth]{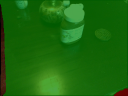}} &
        {\includegraphics[width=0.125\linewidth]{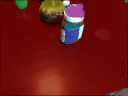}} &
        {\includegraphics[width=0.125\linewidth]{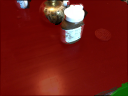}}\\

        {\includegraphics[width=0.125\linewidth]{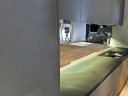}} &  
        {\includegraphics[width=0.125\linewidth]{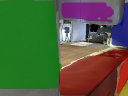}} &  
        {\includegraphics[width=0.125\linewidth]{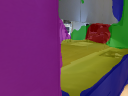}} & 
        {\includegraphics[width=0.125\linewidth]{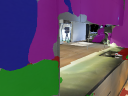}} &
        {\includegraphics[width=0.125\linewidth]{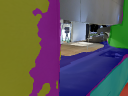}} &
        {\includegraphics[width=0.125\linewidth]{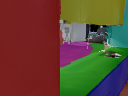}} &
        {\includegraphics[width=0.125\linewidth]{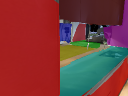}}\\

        {\includegraphics[width=0.125\linewidth]{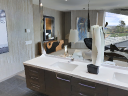}} &  
        {\includegraphics[width=0.125\linewidth]{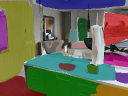}} &  
        {\includegraphics[width=0.125\linewidth]{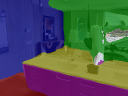}} & 
        {\includegraphics[width=0.125\linewidth]{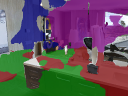}} &
        {\includegraphics[width=0.125\linewidth]{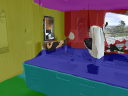}} &
        {\includegraphics[width=0.125\linewidth]{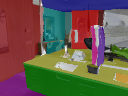}} &
        {\includegraphics[width=0.125\linewidth]{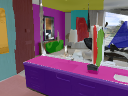}}\\

        {\includegraphics[width=0.125\linewidth]{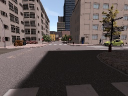}} &  
        {\includegraphics[width=0.125\linewidth]{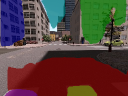}} &  
        {\includegraphics[width=0.125\linewidth]{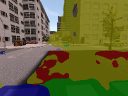}} & 
        {\includegraphics[width=0.125\linewidth]{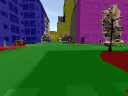}} &
        {\includegraphics[width=0.125\linewidth]{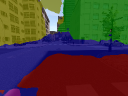}} &
        {\includegraphics[width=0.125\linewidth]{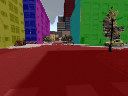}} &
        {\includegraphics[width=0.125\linewidth]{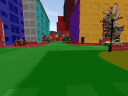}}\\

        {\includegraphics[width=0.125\linewidth]{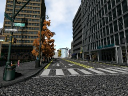}} &  
        {\includegraphics[width=0.125\linewidth]{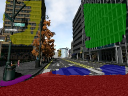}} &  
        {\includegraphics[width=0.125\linewidth]{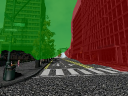}} & 
        {\includegraphics[width=0.125\linewidth]{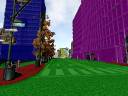}} &
        {\includegraphics[width=0.125\linewidth]{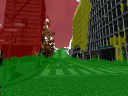}} &
        {\includegraphics[width=0.125\linewidth]{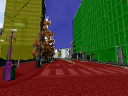}} &
        {\includegraphics[width=0.125\linewidth]{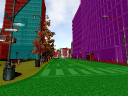}}\\

        Image &  PlaneRCNN~\cite{liu2019planercnn} & PlaneTR~\cite{tan2021planetr} & PlaneRecover~\cite{yang2018recovering} & PlaneRecTR~\cite{shi2023planerectr} & MonoPlane & Groundtruth \\

    \end{tabular}
    \caption{Qualitative results for single-view plane segmentation. We show two samples for ScanNet~\cite{dai2017scannet}~(row 1\&2), Matterport3D~\cite{chang2017matterport3d}~(row 3\&4), and Synthia~\cite{ros2016synthia}~(row 5\&6), respectively, for all methods.}
    \label{fig::singleview}
    \vspace{-10pt}
\end{figure*}
\subsubsection{Image-level Multi-plane Joint Optimization}
Although the proximity-based sequential GC-RANSAC delivers elegant plane reconstruction results, the plane masks may have minor inconsistency with each other due to the sequential binary labeling manner. Thus we apply the densely connected conditional random fields (CRFs)~\cite{krahenbuhl2011efficient} as a post-processing step to jointly optimize multiple plane masks, resulting in more pleasing results. 
We also incorporate the monocular cues inside the CRFs to better model the pairwise potentials, with a formulation similar to  Eq.~\ref{eq::kp},~\ref{eq::kc},~\ref{eq::kn}. 

\subsection{Extension to Sparse-view Images}
\label{sec::multiview}
The proposed single-view pipeline can be easily extended to sparse views. For simplicity, here we use two view inputs to demonstrate (Figure~\ref{fig::pipeline_sparse}). We begin by using the Structure-from-Motion (SfM) to get the camera parameters and sparse 3D points. We then align the monocular depths of multiple images with this sparse reconstruction to mitigate the scale inconsistency. After that, single view plane reconstruction is applied for each image to get the 3D plane proposals. Finally we match these plane proposals between image pairs and fuse the matched planes into global 3D planes.

\noindent\textbf{Camera poses and scale alignment.} We adopt the COLMAP~\cite{schonberger2016structure} to provide camera parameters. 
To improve accuracy, we utilize LoFTR~\cite{sun2021loftr} as the feature extractor and matcher. 
Since depths of different views are predicted independently and have no scale consistency guaranteed, we explicitly align each depth map with the SfM sparse reconstruction by solving a least squares problem. 

\noindent\textbf{Plane matching and fusion.} 
Plane matching between two images is jointly determined by the 2D sparse pixel matching from SfM and the plane re-projection overlap check. 
As for the fusion, we use the average plane as the fused plane of matched pairs. Following~\cite{jin2021planar}, the offsets are averaged and the fused normal $\mathbf{n}$ is solved by maximizing the $\sum_{i}(\mathbf{n^T}\mathbf{n}_i)^2$, where $i=1,2$ for two view pairs. 
\section{EXPERIMENTS}
\label{sec:experiments}

\begin{table*}[]
\centering
\caption{Zero-shot single-view evaluation on indoor datasets NYUv2~\cite{silberman2012indoor} and Matterport3D~\cite{chang2017matterport3d}. ``ScanNet/Matterport3D'' indicates that method is trained on ScanNet~\cite{dai2017scannet} or Matterport3D~\cite{chang2017matterport3d}. ``Omnidata/ZeroDepth'' denotes our versions using Omnidata~\cite{eftekhar2021omnidata} or ZeroDepth~\cite{guizilini2023towards} depth prediction as input. The best and second best results are highlighted by boldface and underline.}
\resizebox{0.7\linewidth}{!}{
\begin{tabular}{lllllll}
\toprule
\centering
\multirow{2}{*}{Method / Metrics} & \multicolumn{3}{c}{Plane Segmentation Metrics} & \multicolumn{3}{c}{Plane Recall (depth)} \\ \cmidrule(r){2-7} 
                                  & VOI~(↓)   & RI~(↑)  & \multicolumn{1}{l}{SC~(↑)}  & @0.05m     & @0.1m   & @0.6m   \\ 
\midrule 
\multicolumn{7}{c}{NYUv2~\cite{silberman2012indoor}} \\
\midrule
PlaneRCNN~(ScanNet)~\cite{liu2019planercnn} & 1.540 & 0.850 & 0.627 & 0.039 & 0.123 & 0.447 \\
PlaneTR~(ScanNet)~\cite{tan2021planetr}  & 1.111 & 0.899 & 0.725 & 0.039  & 0.113 & 0.333 \\ 
PlaneRecTR~(ScanNet)~\cite{shi2023planerectr} & 1.045 & \underline{0.915} & \underline{0.745}  & 0.06 & 0.157 & 0.443\\
MonoPlane~(Omnidata) & \textbf{0.907} & \textbf{0.927} & \textbf{0.770} & \textbf{0.069} & \textbf{0.211} & \textbf{0.497}\\
MonoPlane~(ZeroDepth) & \underline{1.011} & \underline{0.915} & 0.744 & \underline{0.066} & \underline{0.172} & \underline{0.449}\\
\midrule
\multicolumn{7}{c}{Matterport3D~\cite{chang2017matterport3d}} \\
\midrule
PlaneRCNN~(ScanNet)~\cite{liu2019planercnn} & 1.826 & 0.837 & 0.585 &  0.027 & 0.084 & 0.367\\
PlaneTR~(ScanNet)~\cite{tan2021planetr}  & 1.553 & 0.846 & 0.610  & 0.027  & 0.068 & 0.220 \\ 
PlaneRecTR~(ScanNet)~\cite{shi2023planerectr} & 1.419 & 0.875 & 0.655 & 0.045 & 0.108 & 0.330 \\
MonoPlane~(Omnidata) & \textbf{1.135} & \textbf{0.896} & \textbf{0.709} & \textbf{0.074} & \textbf{0.188} & \textbf{0.464}\\
MonoPlane~(ZeroDepth) & \underline{1.191} & \underline{0.890} & \underline{0.693} & \underline{0.059} & \underline{0.155} & \underline{0.417} \\
\bottomrule
\end{tabular}}
\vspace{-10pt}
\label{tab::singleview-merged}
\end{table*}
\subsection{Implementation and Evaluation Details }
We test Omnidata~\cite{eftekhar2021omnidata} and ZeroDepth~\cite{guizilini2023towards} as our pre-trained zero-shot monocular network to predict depths. Omnidata~\cite{eftekhar2021omnidata} is also used to predict surface normals. The hyper-parameters of our method are set as: the distance threshold $\epsilon$ as 0.02, the normal angle threshold $\epsilon_n$ as 10, the smoothness weight $\lambda$ as 0.95. For proximity modeling, the parameters of each similarity term are set as: position term $\alpha^2_p$ as 0.005, image color term $\alpha^2_c$ as 0.1, normal term $\alpha^2_n$ as 5.0. $w_p, w_c, w_n$ are all set to 1. We use the same hyper-parameter setting in all experiments.

For single-view evaluation, we follow the evaluation protocol of PlaneRCNN~\cite{liu2019planercnn}. Three metrics named Variation of Information (VOI), Rand Index (RI), and Segmentation Covering (SC) are used to assess plane segmentation. For planar reconstruction quality, we evaluate per-plane recall under different reconstructed depth gap thresholds.

For sparse-view evaluation, we follow the protocol of SparsePlane~\cite{jin2021planar} except that we filter out the extreme image pairs that have below $10\%$ visual overlap. The plane reconstruction accuracy is evaluated by AP given certain mask IoU, plane normal and offset thresholds.

For both single-view and sparse-view evaluation, to eliminate the scale ambiguity from monocular input, for every method, we align the inferred depth with groundtruth depth by multiplying the median scale factor of the groundtruth depth~\wrt~the inferred depth map before evaluation. 

\noindent\textbf{Running time.} On average, for a single 192x256 image, the proposed RANSAC takes 0.67 seconds, and the dense CRFs takes 0.22 seconds. For two-view images, the LoFTR
matching and the COLMAP SfM in total take 1.10 seconds.

\subsection{Single-view Plane Reconstruction Evaluation}

\begin{table}[]
\centering
\caption{Zero-shot single-view evaluation on outdoor dataset Synthia~\cite{ros2016synthia}.}
\resizebox{1.0\linewidth}{!}{
\begin{tabular}{lllllll}
\toprule
\centering
\multirow{2}{*}{Method / Metrics} & \multicolumn{3}{c}{Plane Segmentation Metrics} & \multicolumn{3}{c}{Plane Recall (depth)} \\ \cmidrule(r){2-7} 
                                  & VOI~(↓)   & RI~(↑)  & \multicolumn{1}{l}{SC~(↑)}  & @1m     & @3m   & @10m    \\ 
                                  
\midrule
PlaneRCNN~(ScanNet)~\cite{liu2019planercnn} & 1.881 & 0.715 & 0.540 & \underline{0.092} & 0.121 & 0.143\\
PlaneTR~(ScanNet)~\cite{tan2021planetr}  & 1.254 & 0.849 & 0.689 &  0.027 & 0.088 & 0.133\\
PlaneRecTR~(ScanNet)~\cite{shi2023planerectr} & 1.261 & 0.852 & 0.698 & 0.027 & 0.119 & 0.165\\
MonoPlane~(Omnidata) & \underline{1.060} & \underline{0.898} & \underline{0.750} & 0.089 & \underline{0.178} & \underline{0.223} \\
MonoPlane~(ZeroDepth) & \textbf{1.032} & \textbf{0.899} & \textbf{0.756} & \textbf{0.119} & \textbf{0.179} & \textbf{0.243}\\
\bottomrule
\end{tabular}}
\vspace{-10pt}
\label{tab::singleview-synthia}
\end{table}

We report the single-view plane reconstruction results on two zero-shot indoor datasets NYUv2~\cite{silberman2012indoor},  Matterport3D~\cite{chang2017matterport3d} and an outdoor dataset Synthia~\cite{ros2016synthia}. Baseline methods include PlaneRCNN~\cite{liu2019planercnn}, PlaneTR~\cite{tan2021planetr} and PlaneRecTR~\cite{shi2023planerectr} which are trained on ScanNet~\cite{dai2017scannet}. As shown in Tab.~\ref{tab::singleview-merged}, For both NYUv2 and Matterport3D which is unseen during training for all methods, our proposed method MonoPlane significantly outperforms the baselines on all metrics. On Synthia (i.e. Tab.~\ref{tab::singleview-synthia}), MonoPlane also clearly outperforms the remaining methods which transfer from indoor scenes. 
See Fig.~\ref{fig::singleview} for qualitative examples. In general, none of those learning-based methods can adapt well to diverse inputs. Contrastively, our method detects and reconstructs planes in a robust manner, and generalizes well to different scenes, which is the main merit and motivation of this work.


\subsection{Sparse-view Plane Reconstruction Evaluation}

\begin{figure*}[tb]
    \centering
    \scriptsize
    \setlength{\tabcolsep}{2pt}
    \begin{tabular}{ccccc}

    \centering
        {\includegraphics[width=0.13\linewidth]{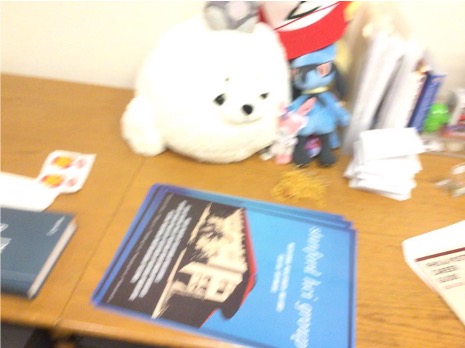}} &  
        {\includegraphics[width=0.13\linewidth]{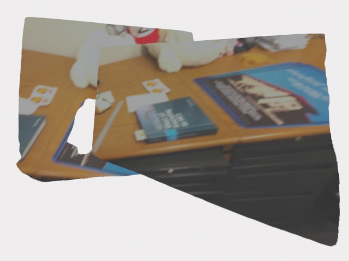}} &  
        {\includegraphics[width=0.13\linewidth]{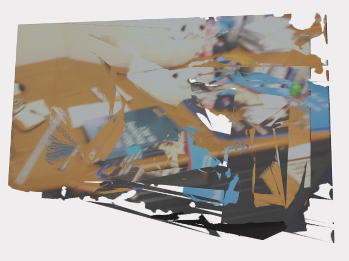}} & 
        {\includegraphics[width=0.13\linewidth]{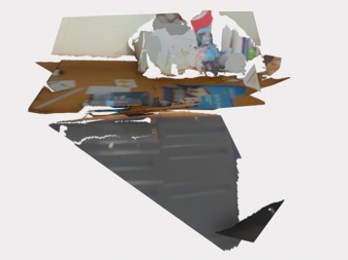}} &
        {\includegraphics[width=0.13\linewidth]{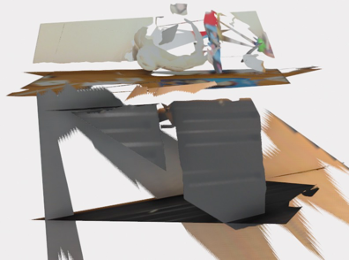}}\\

        {\includegraphics[width=0.13\linewidth]{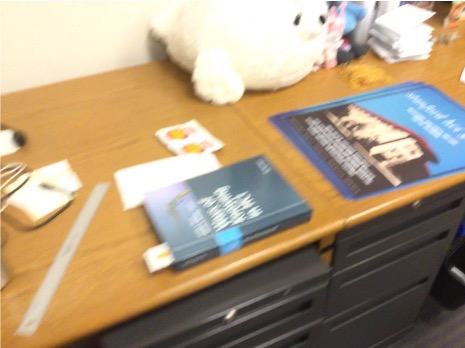}} &  
        {\includegraphics[width=0.13\linewidth]{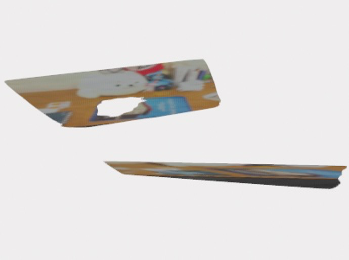}} &  
        {\includegraphics[width=0.13\linewidth]{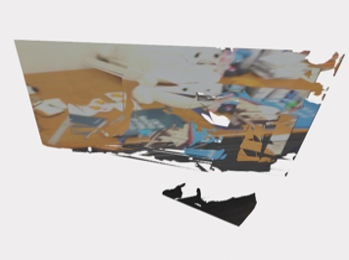}} & 
        {\includegraphics[width=0.13\linewidth]{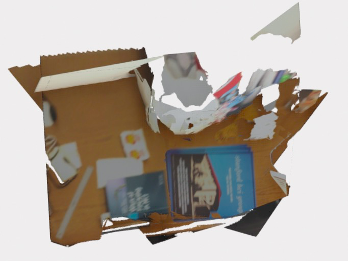}} &
        {\includegraphics[width=0.13\linewidth]{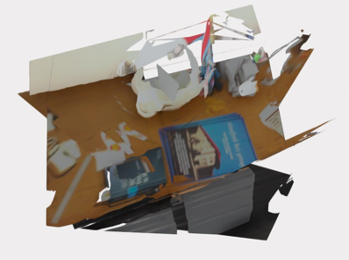}}\\
        \midrule

        {\includegraphics[width=0.13\linewidth]{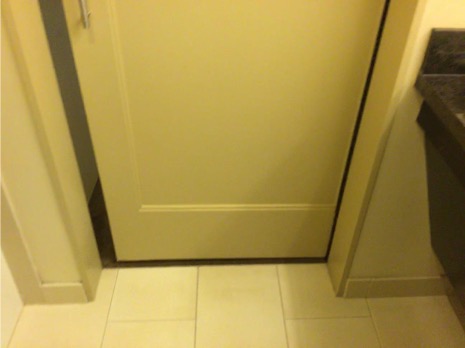}} &  
        {\includegraphics[width=0.13\linewidth]{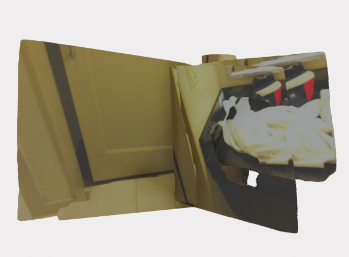}} &  
        {\includegraphics[width=0.13\linewidth]{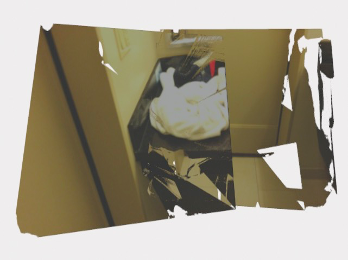}} & 
        {\includegraphics[width=0.13\linewidth]{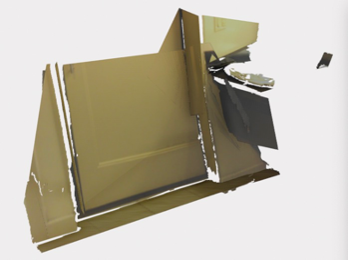}} &
        {\includegraphics[width=0.13\linewidth]{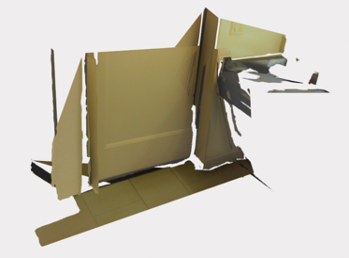}}\\

        {\includegraphics[width=0.13\linewidth]{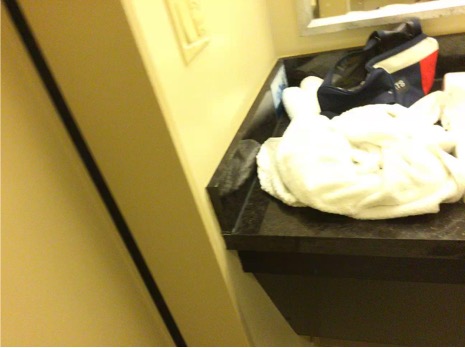}} &  
        {\includegraphics[width=0.13\linewidth]{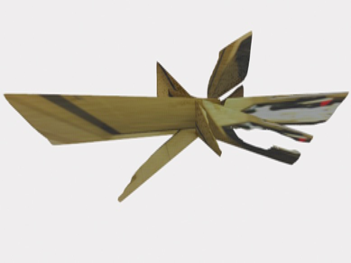}} &  
        {\includegraphics[width=0.13\linewidth]{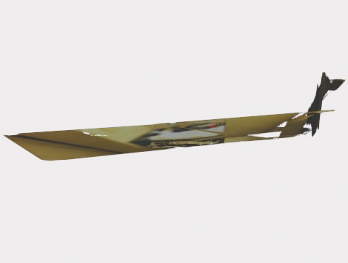}} & 
        {\includegraphics[width=0.13\linewidth]{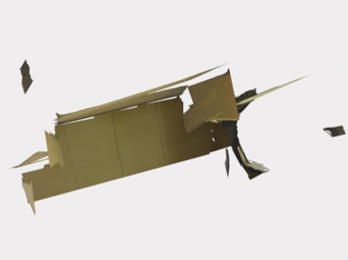}} &
        {\includegraphics[width=0.13\linewidth]{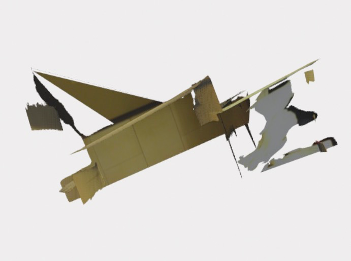}}\\

        Image-pairs &  PlaneFormers~\cite{agarwala2022planeformers} & NopeSAC~\cite{tan2022nope} & MonoPlane & Groundtruth \\

    \end{tabular}
    \caption{Reconstructed planar mesh from two views of ScanNet~\cite{dai2017scannet} dataset. We show a frontal view~($1st$ and $3rd$ rows) and a bird-eye view~($2nd$ and $4th$ rows) for each case respectively.}
    \label{fig::twoview-scannet}
\end{figure*}
\begin{table}[]
\centering
\caption{Zero-shot two-view evaluation on ScanNet~\cite{dai2017scannet} dataset.}
\resizebox{1.0\linewidth}{!}{
\begin{tabular}{lcccc}
\toprule
\centering
\multirow{2}{*}{Method / Metrics} & \multicolumn{4}{c}{Offset $\leq$ 1m, Normal $\leq$ 30°} \\ \cmidrule(r){2-5} 
                                  & All   & -Offset  & -Normal  & -Offset-Normal   \\ 
                                  
\midrule 
PlaneFormers~(Matterport3D)~\cite{agarwala2022planeformers}  & 2.07 & 2.77  & 5.04  & 8.69 \\
NopeSAC~(Matterport3D)~\cite{tan2022nope}  & 2.47  & 3.17   & 5.13   & 7.09 \\ 
MonoPlane~(Omnidata)  & \textbf{9.15}  & \textbf{15.74}   & \textbf{12.21}   & \textbf{21.16} \\
\bottomrule
\end{tabular}
}
\vspace{-10pt}
\label{tab::twoview-AP}
\end{table}

Two-view plane reconstruction quality is evaluated on ScanNet dataset, compared with two SOTA methods PlaneFormers~\cite{agarwala2022planeformers} and NopeSAC~\cite{tan2022nope} which are both trained on Matterport3D. The results are summarized in Tab.~\ref{tab::twoview-AP}. MonoPlane 
significantly outperforms these counterparts. As shown in Fig.~\ref{fig::twoview-scannet}, PlaneFormers~\cite{agarwala2022planeformers} and NopeSAC~\cite{tan2022nope} fail to recover the major planar surface when exposed to a new environment with different distribution from the training data. Conversely, our approach showcases superior robustness on these challenging scenarios, clearly demonstrating its scalability when extended to sparse-view input.

\subsection{In-the-wild Data}
\begin{figure*}[tb]
    \centering
    \scriptsize
    \setlength{\tabcolsep}{2pt}
    \begin{tabular}{cccccc}

    \centering
        {\includegraphics[width=0.14\linewidth]{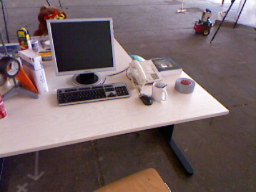}} &  
        {\includegraphics[width=0.14\linewidth]{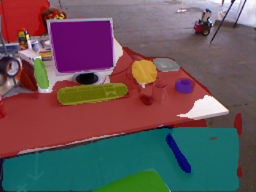}} &  
        {\includegraphics[width=0.14\linewidth]{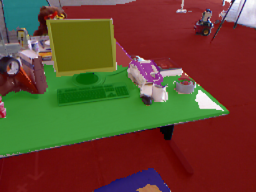}} &

        {\includegraphics[width=0.14\linewidth]{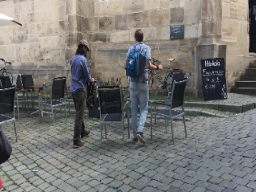}} &  
        {\includegraphics[width=0.14\linewidth]{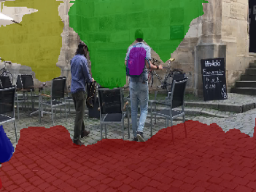}} &  
        {\includegraphics[width=0.14\linewidth]{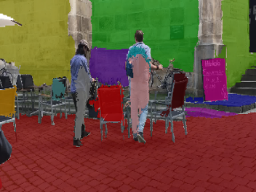}} \\

        {\includegraphics[width=0.14\linewidth]{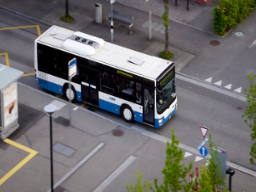}} &  
        {\includegraphics[width=0.14\linewidth]{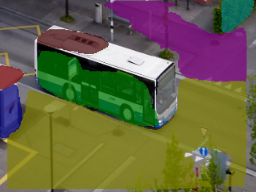}} &  
        {\includegraphics[width=0.14\linewidth]{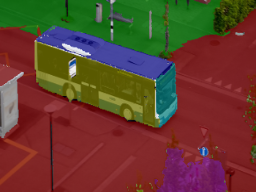}} &

        {\includegraphics[width=0.14\linewidth]{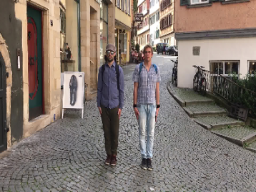}} &  
        {\includegraphics[width=0.14\linewidth]{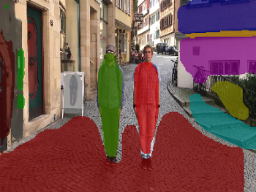}} &  
        {\includegraphics[width=0.14\linewidth]{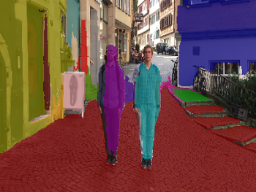}} \\



        Image &  PlaneRCNN~\cite{liu2019planercnn} & MonoPlane & Image &  PlaneRCNN~\cite{liu2019planercnn} & MonoPlane 

    \end{tabular}
    \caption{Plane segmentation results on in-the-wild data from TUM-RGBD~\cite{sturm2012benchmark}, 3DPW~\cite{vonMarcard2018} and DAVIS~\cite{perazzi2016benchmark}.}
    \label{fig::wild}
    \vspace{-10pt}
\end{figure*}
We further test our framework on in-the-wild data from several publicly available datasets, including 3DPW~\cite{vonMarcard2018}, KITTI~\cite{geiger2013vision}, TUM-RGBD~\cite{sturm2012benchmark}, DAVIS~\cite{perazzi2016benchmark}, etc. Fig.~\ref{fig::wild} shows the qualitative comparison between our method and PlaneRCNN~\cite{liu2019planercnn}. Our method MonoPlane transfers well to the in-the-wild scenes and clearly outperforms the learning-based counterpart. To further demonstrate the feasibility of our method in practical applications such as robotics environment, we also apply our system on a real-world scenario using a mobile robot. This real-world application and more qualitative results on single-view and sparse-view reconstruction of in-the-wild data are shown in the supplementary video.

\subsection{Ablation Studies}
We conduct several ablation experiments of single-view plane reconstruction on ScanNet dataset. As shown in Tab.~\ref{tab::ablation}, our method surpasses the sequential RANSAC and Graph-cut RANSAC by a large margin, validating the effectiveness of our system design. Figure~\ref{fig::ablation} showcases the qualitative comparison of gradually adding our proposed components. 
Sequential-RANSAC produces messy plane fragments with incomplete masks. While GC-RANSAC greatly improves the results thanks to its coherence modeling, our method outperforms both of the two baselines and delivers more accurate and complete plane reconstructions. 
We also verified the necessity of the dense CRF module and conducted an ablation study. As illustrated, the dense CRF can merge or remove some small fuzzy planes, and make the plane segmentation more coherent around image boundaries.

\begin{table}[]
\centering
\caption{Ablation studies on ScanNet~\cite{dai2017scannet} dataset.}
\resizebox{1.0\linewidth}{!}{
\begin{tabular}{lcccccc}
\toprule
\centering
\multirow{2}{*}{Method / Metrics} & \multicolumn{3}{c}{Plane Segmentation Metrics} & \multicolumn{3}{c}{Plane Recall (depth)} \\ \cmidrule(r){2-7} 
                                  & VOI~(↓)   & RI~(↑)  & \multicolumn{1}{l}{SC~(↑)}  & @0.05m     & @0.1m   & @0.6m    \\ 

\midrule
Seq-RANSAC~\cite{fischler1981random} & 2.543 & 0.677 & 0.407 &  0.108 & 0.189 & 0.247 \\
Seq-RANSAC + CRFs &  1.438 & 0.787 & 0.637 & 0.152 & 0.284 & 0.400 \\
GC-RANSAC~\cite{barath2018graph}  & 1.290 & 0.852 & 0.717 &  0.144 & 0.288 & 0.417 \\
GC-RANSAC + CRFs  &  1.028 & 0.872 & 0.758 & 0.157 & 0.322 & 0.492 \\
Ours w/o CRFs &  \underline{1.004} &  \underline{0.906} & \underline{0.787} &  \underline{0.205} & \underline{0.400} & \underline{0.612} \\
Ours  &  \textbf{0.910} & \textbf{0.912} & \textbf{0.798} & \textbf{0.214} & \textbf{0.424} & \textbf{0.662} \\
\bottomrule
\end{tabular}}
\vspace{-15pt}
\label{tab::ablation}
\end{table}
\section{CONCLUSION}
\label{sec:conclusion}

In this paper, we present a novel framework named MonoPlane, to detect and reconstruct 3D planes from a single RGB image. The key idea is exploiting monocular geometric cues (depth and surface normal) to guide the point proximity based graph-cut RANSAC for 3D plane fitting. We further extend our proposed single-view approach to sparse-view plane reconstruction.
MonoPlane achieves superior generalizability across different datasets, making it suitable for in-the-wild plane reconstruction in various downstream applications in robotics or augmented reality.

\noindent{\it Limitations and future work.} Our method relies on the pretrained monocular geometry networks. While delivering remarkable transferability, it may fail when geometric predictions have great errors. In the future, we aim to extend our point-to-point proximity with more similarity measurements such as deep image features. Off-the-shelf segmentation from large models such as SAM~\cite{kirillov2023segment} can also be used to detect planes and accelerate RANSAC process. We also plan to incorporate multi-view optimizations such as plane-camera bundle-adjustment into our sparse-view pipeline.

\noindent{\it Acknowledgements.} We sincerely thank the reviewers for their constructive feedback. We are grateful to Shaohui Liu for helpful discussions.
\section{APPENDIX}
\label{sec:appendix}

\subsection{More Details of Pipeline}
\paragraph{Dense CRFs.} After running our proposed RANSAC algorithm on 3D point cloud to get plane instance masks and parameters, we further utilize the densely-connected conditional random fields (DCRFs) to improve the masks at image level. Specifically, the plane masks are first projected onto the image as a multi-label segmentation map, then we define and solve the DCRFs based on~\cite{krahenbuhl2011efficient}. Different from the neighborhood graph defined in Graph-cut RANSAC~\cite{barath2018graph}, DCRFs operate on the densely-connected graph where every pixel is connected to each other. Efficient solver is proposed in \cite{krahenbuhl2011efficient} with gaussian edge potentials. Unary potential $\psi_u$ and pairwise potential $\psi_p$ are defined as the optimization objective $E$: 
\begin{equation}
    E(x) = \sum_i \psi_u(x_i) + \sum_{i<j} \psi_p(x_i, x_j)
\label{eq::crf}
\end{equation}
where $x$ denotes the image pixel. As the common practice in segmentation, the unary potential is defined as the difference between the new labels and the original labels (from point cloud). The pairwise potential is defined as a set of pairwise gaussian kernels. Here we propose smoothness, appearance, depth and surface normal kernels to optimize the label map.
\begin{align}
\label{eq::kp}
    & k_s = \exp \left( -\frac{|p_i - p_j|^2}{2\theta^2_s} \right) \\
    \label{eq::kc}
    & k_a = \exp \left( -\frac{|p_i - p_j|^2}{2\theta^2_s}  - \frac{|I_i - I_j|^2}{2\theta^2_a}\right) \\
    \label{eq::kn}
    & k_d = \exp \left( -\frac{|p_i - p_j|^2}{2\theta^2_s}  - \frac{|N_i - N_j|^2}{2\theta^2_d}\right) \\
    & k_n = \exp \left( -\frac{|p_i - p_j|^2}{2\theta^2_s}  - \frac{|D_i - D_j|^2}{2\theta^2_n}\right) \\
    & \psi_p(x_i, x_j) = \mu(x_i, x_j)\sum_{m=1}^{K}w^{(m)}k^{(m)}(\bold{f}_i, \bold{f}_j)
\end{align}
where $\mu(x_i, x_j) = \left[x_i \ne x_j\right]$ is the Potts model, and $\bold{f}_i$ denotes the per-pixel features such as position $p_i$, image color vector $I_i$, depth $D_i$ and normal vector $N_i$. Depths and normals are from monocular predictions. In all our experiments, the weights $w^{(m)}$ is set to one, and the deviation parameters $\left[ \theta_s, \theta_a, \theta_d, \theta_n\right]$ are set as $[10, 10, 0.1, 0.01]$. 

After applying dense CRFs, we further run connected component checking to split non-connected planar regions and remove small regions, e.g. less than 300 pixels, to erase small plane fragments.

\begin{figure*}[tb]
    \centering
    \scriptsize
    \setlength{\tabcolsep}{2pt}
    \begin{tabular}{ccccccc}

    \centering
        {\includegraphics[width=0.125\linewidth]{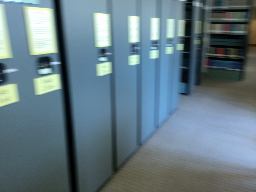}} &  
        {\includegraphics[width=0.125\linewidth]{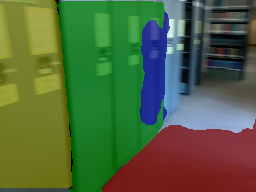}} &  
        {\includegraphics[width=0.125\linewidth]{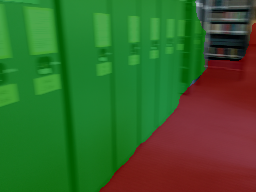}} & 
        {\includegraphics[width=0.125\linewidth]{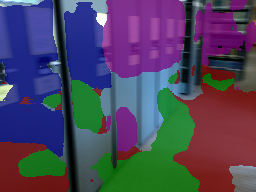}} &
        {\includegraphics[width=0.125\linewidth]{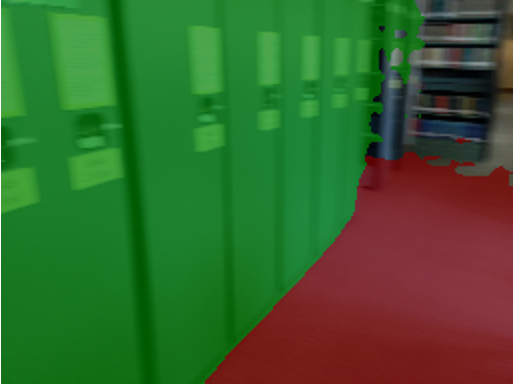}} &
        {\includegraphics[width=0.125\linewidth]{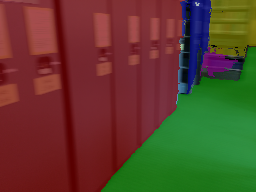}} &
        {\includegraphics[width=0.125\linewidth]{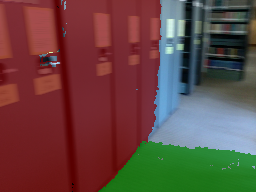}}\\

        {\includegraphics[width=0.125\linewidth]{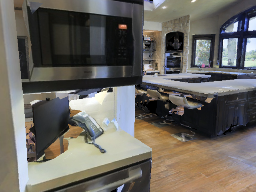}} &  
        {\includegraphics[width=0.125\linewidth]{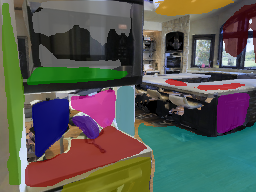}} &  
        {\includegraphics[width=0.125\linewidth]{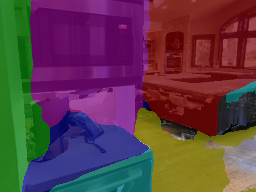}} & 
        {\includegraphics[width=0.125\linewidth]{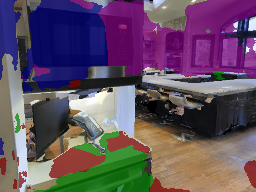}} &
        {\includegraphics[width=0.125\linewidth]{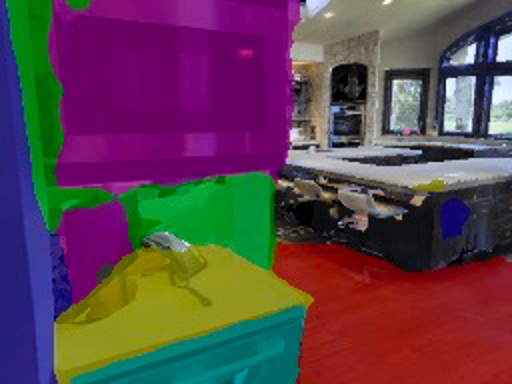}} &
        {\includegraphics[width=0.125\linewidth]{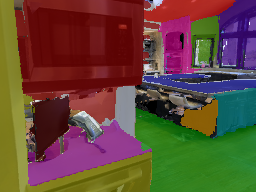}} &
        {\includegraphics[width=0.125\linewidth]{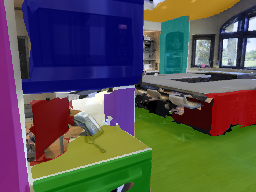}}\\

        {\includegraphics[width=0.125\linewidth]{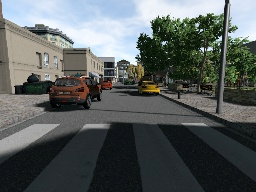}} &  
        {\includegraphics[width=0.125\linewidth]{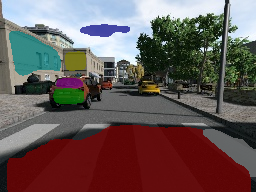}} &  
        {\includegraphics[width=0.125\linewidth]{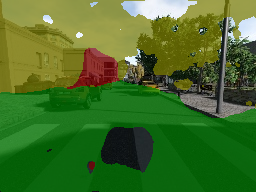}} & 
        {\includegraphics[width=0.125\linewidth]{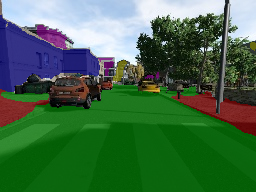}} &
        {\includegraphics[width=0.125\linewidth]{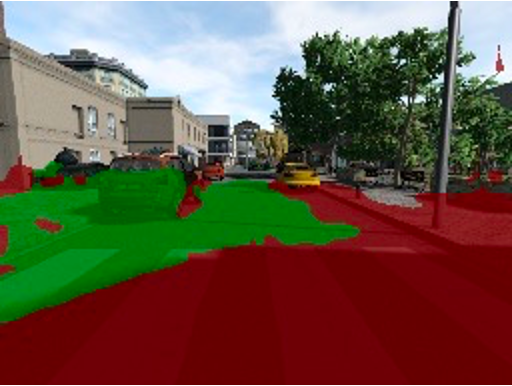}} &
        {\includegraphics[width=0.125\linewidth]{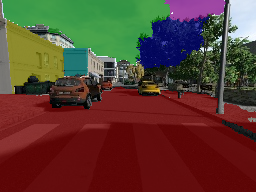}} &
        {\includegraphics[width=0.125\linewidth]{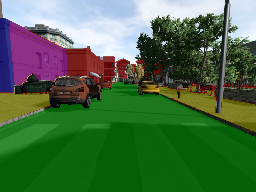}}\\

        Image &  PlaneRCNN & PlaneTR & PlaneRecover & PlaneRecTR & Ours & Groundtruth \\

    \end{tabular}
    \caption{More single-view plane segmentation qualitative results on ScanNet, Matterport3D and Synthia datasets.}
    \label{fig::supp_singleview}
\end{figure*}

\paragraph{Sparse-view extension.} The sparse-view pipeline is easily extended from single-view. To acquire camera parameters, we run the structure-from-motion algorithm COLMAP~\cite{schonberger2016structure}. Camera intrinsics are fixed and input to the SfM if available, otherwise estimated together with extrinsics by SfM. 
The SfM outputs for image $I_i$ are camera extrinsic ${R_i, t_i}$, shared intrinsic $K$, and sparse depths maps $D^s_i$ by projecting sparse 3d points onto each image. Monocular depths are predicted separately for each image, thus have no scale consistency guaranteed, which hinders the global 3d plane reconstruction. We solve this scale ambiguity by aligning each monocular depth onto the SfM sparse reconstruction. Specifically, given monocular depth $D^m_i$ for each image, we solve the least square problem $|M^s_i * (D^s_i - s * D^m_i)|^2$ to get the scaling factor $s$, where $M^s_i = (D^s_i > 0)$ is the valid mask of sparse SfM depths. Scaled monocular depth $s*D^m_i$ is then used for the following steps. 

For the plane matching, we use the combination of pixel matching and plane re-projection overlap check to determine the matching. Specifically, if two planes of different images contain sufficient sparse keypoint matches (from SfM), e.g. more than 10 pixel matches, and pass the keypoint mutual check, they will be viewed as matched planes. The keypoint mutual check is to ensure that most of the keypoints (more than 60\%) of the current plane are matched with the keypoints of the candidate matched plane, instead of other planes. Since sparse pixel matches can not uniformly cover the whole image, we thus use plane re-projection to find additional plane matches. The plane in image $i$ is re-projected onto image $j$ using estimated plane parameters and camera poses. If the re-projected plane has a large intersection-of-union value with a plane in their visible regions, we define these two planes as matched.

For the plane fusion, we use the averaged plane parameter as the fused one. From two planes with parameters $\mathbf{n_1}, \mathbf{o_1}, \mathbf{n_2}, \mathbf{o_2}$ in global coordinate, the averaged plane is defined as the plane with the fused normal $\mathbf{n}$ solved by maximizing the $\sum_{i}(\mathbf{n^T}\mathbf{n}_i)^2$, where $i=1,2$ for two view pairs. The averaged offset is $\mathbf{o} = \frac{\mathbf{o_1} + \mathbf{o_2}}{2}$.

\paragraph{Spare view running time analysis.} We report the running time of each component for the sparse view pipeline here. Our experiments are conducted with Intel Xeon(R) Platinum 8375C CPUs and a Nvidia A100 GPU. We calculate the averaged running time for each module on the ScanNet dataset. For two-view images with 192x256 resolution, the LoFTR matching and the COLMAP SfM in total take 1.10 seconds. Our proposed single view pipeline costs 0.89 seconds for each image. And finally the plane matching and fusion take 0.05 seconds. Further parallelization can be used to accelerate the RANSAC process.

\begin{figure}[tb]
    \centering
    \scriptsize
    \setlength{\tabcolsep}{2pt}
    \begin{tabular}{ccc}

    \centering
        {\includegraphics[width=0.30\linewidth]{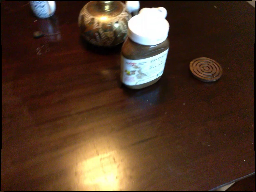}} &  
        {\includegraphics[width=0.30\linewidth]{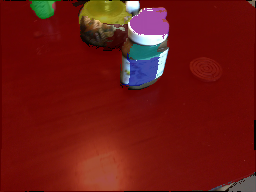}} &  
        {\includegraphics[width=0.30\linewidth]{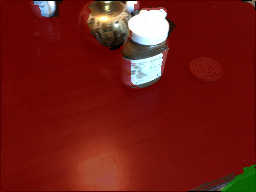}} \\

        Image & Reconstruction &  Groundtruth

    \end{tabular}
    \caption{An example which demonstrates the ambiguous annotations in ScanNet~\cite{dai2017scannet} dataset.}
    \label{fig::supp_nonplanar_gt_eval}
\end{figure}
\begin{table}[]
\centering
\resizebox{1.0\linewidth}{!}{
\begin{tabular}{lllllll}
\toprule
\centering
\multirow{2}{*}{Method / Metrics} & \multicolumn{3}{c}{Translation} & \multicolumn{3}{c}{Rotation} \\ \cmidrule(r){2-7} 
                                  & Med.~(↓)   & Mean~(↓)  & \multicolumn{1}{c}{$\leq$ 1m}  & Med.~(↓)     & Mean~(↓)   & $\leq$ 30°   \\ 
                                  
\midrule 
\multicolumn{7}{c}{ScanNet~\cite{dai2017scannet}} \\
\midrule
PlaneFormers~\cite{agarwala2022planeformers}~(Matterport3D)                      & 0.62     & 0.94    & 70.86 & 39.36     & 50.76    & 37.41 \\ 
NopeSAC~\cite{tan2022nope}~(Matterport3D)                           & 0.59     & 0.90    & 70.94 & 28.42     & 39.67    & 52.41 \\
Ours                              & \textbf{0.21}     & \textbf{0.56}    & \textbf{82.86} & \textbf{10.75}    & \textbf{26.64}    & \textbf{68.80} \\ 
\bottomrule
\end{tabular}
}
\caption{Two-view relative camera pose evaluation on ScanNet~\cite{dai2017scannet}.}

\label{tab::twoview-camera}
\end{table}

\begin{table*}[]
\centering
\resizebox{0.8\linewidth}{!}{
\begin{tabular}{lllllll}
\toprule
\centering
\multirow{2}{*}{Method / Metrics} & \multicolumn{3}{c}{Plane Segmentation Metrics} & \multicolumn{3}{c}{Plane Recall (depth)} \\ \cmidrule(r){2-7} 
                                  & VOI~(↓)   & RI~(↑)  & \multicolumn{1}{l}{SC~(↑)}  & @0.05m     & @0.1m   & @0.6m   \\ 
\midrule 
\multicolumn{7}{c}{NYUv2~\cite{silberman2012indoor}} \\
\midrule
PlaneTR~(ScanNet)~\cite{tan2021planetr} & 1.111 & 0.899 & 0.725  & 0.021 & 0.063 & 0.283 \\ 
PlaneRCNN~(ScanNet)~\cite{liu2019planercnn} & 1.540 & 0.850 & 0.627 & 0.03 & \textbf{0.098} & \textbf{0.434}\\
PlaneRecTR~(ScanNet)~\cite{shi2023planerectr} & \underline{1.045} & \textbf{0.915} & \textbf{0.745} & \textbf{0.033} & \textbf{0.098} & 0.403\\
Ours~(ZeroDepth) & \textbf{1.011} & \textbf{0.915} & \underline{0.744} & \underline{0.029} & \underline{0.08} & \underline{0.428}\\
\midrule
\multicolumn{7}{c}{Matterport3D~\cite{chang2017matterport3d}} \\
\midrule
PlaneTR~(ScanNet)~\cite{tan2021planetr} & 1.559 & 0.846 & 0.609  & 0.009  & 0.024 & 0.143 \\
PlaneRCNN~(ScanNet)~\cite{liu2019planercnn} & 1.826 & 0.837 & 0.585 &  0.02 & 0.058 & 0.309\\
PlaneRecTR~(ScanNet)~\cite{shi2023planerectr} & 1.419 & 0.875 & 0.655 & 0.023 & 0.055 & 0.262\\
Ours~(ZeroDepth) & \textbf{1.186} & \textbf{0.886} & \textbf{0.691} & \textbf{0.033} & \textbf{0.084} & \textbf{0.372}\\
\midrule
\multirow{2}{*}{Method / Metrics} & \multicolumn{3}{c}{Plane Segmentation Metrics} & \multicolumn{3}{c}{Plane Recall (depth)} \\ \cmidrule(r){2-7} 
                                  & VOI~(↓)   & RI~(↑)  & \multicolumn{1}{l}{SC~(↑)}  & @1m     & @3m   & @10m    \\ 
\midrule
\multicolumn{7}{c}{Synthia~\cite{ros2016synthia}} \\
\midrule
PlaneTR~(ScanNet)~\cite{tan2021planetr}  & 1.254 & 0.849 & 0.689 & 0.0 & 0.005 & 0.103\\
PlaneRCNN~(ScanNet)~\cite{liu2019planercnn} & 1.881 & 0.715 & 0.540 & 0.015 & 0.104 & 0.14\\
PlaneRecTR~(ScanNet)~\cite{shi2023planerectr} & 1.261 & 0.852 & 0.698 & 0.0 & 0.021 & 0.145\\
Ours~(ZeroDepth) & \textbf{1.032} & \textbf{0.899} & \textbf{0.756} & \textbf{0.088} & \textbf{0.162} & \textbf{0.232}\\
\bottomrule
\end{tabular}}
\caption{Zero-shot evaluation on plane recall~\wrt metric depth on NYUv2~\cite{silberman2012indoor}, Matterport3D~\cite{chang2017matterport3d} and Synthia~\cite{ros2016synthia} datasets.}
\label{tab::metric_depth}
\end{table*}
\begin{table}[]
\centering
\caption{Applying dense CRFs on top of each learning-based baseline methods on ScanNet~\cite{dai2017scannet} dataset. Plane recall is evaluated under scale-aligned depth prediction.}
\resizebox{0.95\linewidth}{!}{
\begin{tabular}{lllllll}
\toprule
\centering
\multirow{2}{*}{Method / Metrics} & \multicolumn{3}{c}{Plane Segmentation Metrics} & \multicolumn{3}{c}{Plane Recall (depth)} \\ \cmidrule(r){2-7} 
                                  & VOI~(↓)   & RI~(↑)  & \multicolumn{1}{l}{SC~(↑)}  & @0.05m     & @0.1m   & @0.6m   \\ 
\midrule 
\multicolumn{7}{c}{ScanNet~\cite{dai2017scannet}} \\
\midrule
PlaneRCNN~(ScanNet)~\cite{liu2019planercnn} & 1.297 & 0.852 & 0.699 &  0.188 & 0.388 & 0.668\\
PlaneTR~(ScanNet)~\cite{tan2021planetr} & 0.761 & 0.923  & 0.837 & 0.272 & 0.455 & 0.614 \\
PlaneRecTR~(ScanNet)~\cite{shi2023planerectr} & 0.652 & 0.943 & 0.865 & 0.345 & 0.548 & 0.753 \\ 
\midrule
PlaneRCNN~(ScanNet, CRF)~\cite{liu2019planercnn} & 1.157 & 0.864 & 0.726 &  0.161 & 0.381 & 0.642\\
PlaneTR~(ScanNet, CRF)~\cite{tan2021planetr} & 0.761 & 0.923  & 0.837 & 0.312 & 0.47 & 0.599 \\
PlaneRecTR~(ScanNet, CRF)~\cite{shi2023planerectr} & 0.682 & 0.938 & 0.859 & 0.377 & 0.567 & 0.732 \\ 

\bottomrule
\end{tabular}}

\label{tab::supp_crf}
\end{table}
\subsection{More Details of Evaluation}

\paragraph{Depth alignment} For single-view evaluation, to eliminate the effect of depth scale ambiguity, we follow~\cite{ranftl2020towards} to align the scale of predicted planar depth map with groundtruth before evaluation for every approach. For indoor datasets~\cite{dai2017scannet, chang2017matterport3d} we compute the scale for alignment using pixels whose groundtruth depths are below~$10m$ for alignment. For outdoor dataset~\cite{ros2016synthia}, the threshold is set to~$30m$. For sparse-view evaluation, we align their relative pose by scaling its translation with the norm of groundtruth translation before evaluation.

\begin{figure}[tb]
    \centering
    \scriptsize
    \setlength{\tabcolsep}{2pt}
    \begin{tabular}{ccc}

    \centering
        {\includegraphics[width=0.25\linewidth]{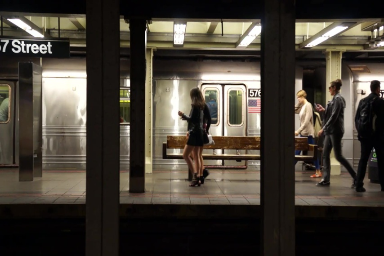}} &  
        {\includegraphics[width=0.25\linewidth]{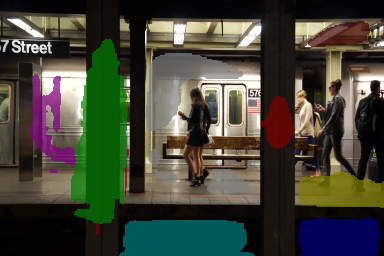}} &  
        {\includegraphics[width=0.25\linewidth]{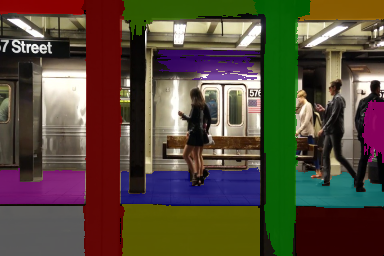}} \\

        {\includegraphics[width=0.25\linewidth]{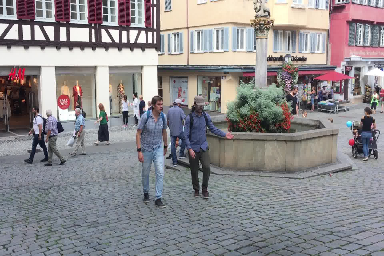}} &  
        {\includegraphics[width=0.25\linewidth]{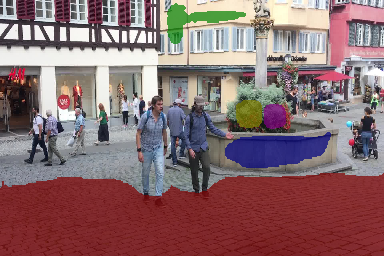}} &  
        {\includegraphics[width=0.25\linewidth]{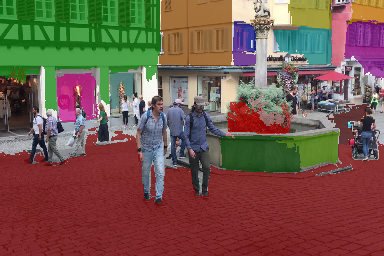}} \\

        {\includegraphics[width=0.25\linewidth]{figures/visualizations/supp_singleviewwild_3dpw_davis/3-1.png}} &  
        {\includegraphics[width=0.25\linewidth]{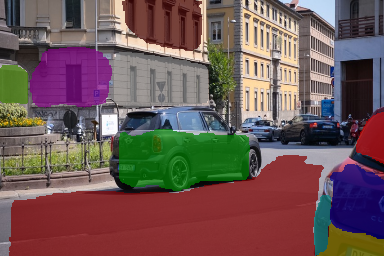}} &  
        {\includegraphics[width=0.25\linewidth]{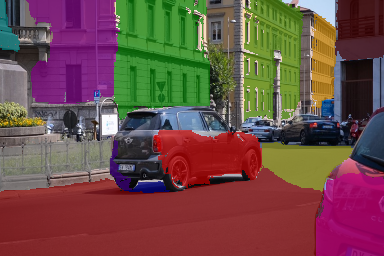}} \\

        Image &  PlaneRCNN & Ours

    \end{tabular}
    \caption{Plane segmentation results on wild~(3DPW \& DAVIS) scenes.}
    \label{fig::supp_singleviewwild_3dpw_davis}
\end{figure}
\begin{figure}[tb]
    \centering
    \scriptsize
    \setlength{\tabcolsep}{2pt}
    \begin{tabular}{ccc}

    \centering
        {\includegraphics[width=0.28\linewidth]{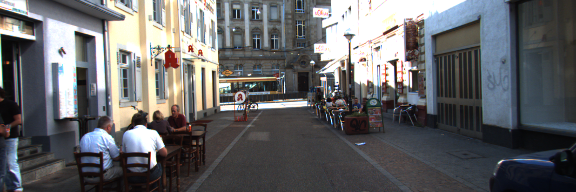}} &  
        {\includegraphics[width=0.28\linewidth]{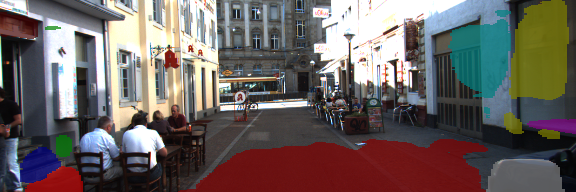}} &  
        {\includegraphics[width=0.28\linewidth]{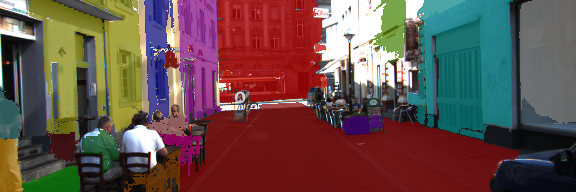}} \\

        {\includegraphics[width=0.28\linewidth]{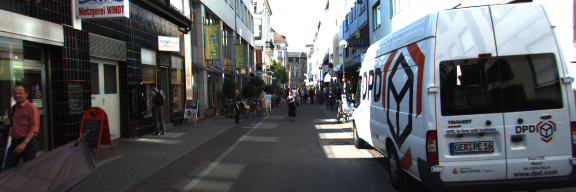}} &  
        {\includegraphics[width=0.28\linewidth]{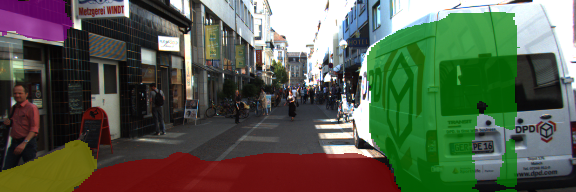}} &  
        {\includegraphics[width=0.28\linewidth]{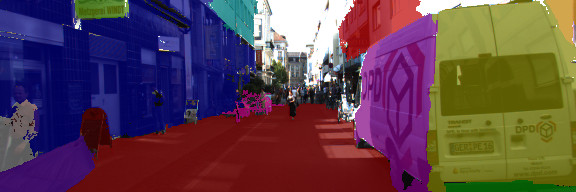}} \\

        {\includegraphics[width=0.28\linewidth]{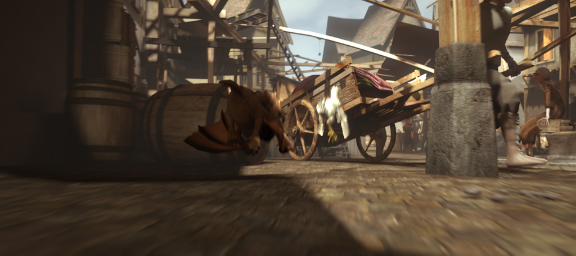}} &  
        {\includegraphics[width=0.28\linewidth]{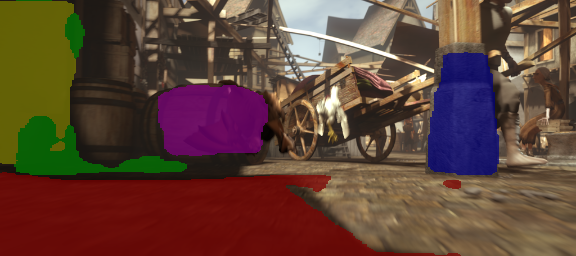}} &  
        {\includegraphics[width=0.28\linewidth]{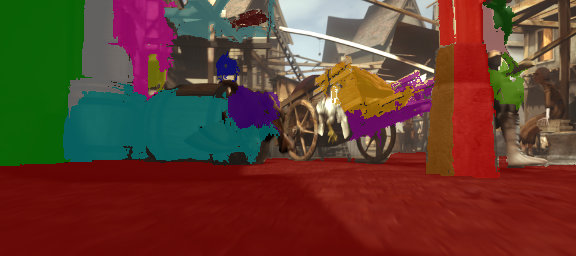}} \\

        Image &  PlaneRCNN & Ours

    \end{tabular}
    \caption{Plane segmentation results on wild~(KITTI \& Sintel) scenes.}
    \label{fig::supp_singleviewwild_sintel_kitti}
\end{figure}

\paragraph{Plane reconstruction evaluation from a geometric perspective} The groundtruth planar annotations used in current work~\cite{liu2019planercnn, jin2021planar} come from the plane-fitting outcomes on top of the groundtruth mesh of selected objects by their semantic categories. Some objects whose semantics are identified as non-planes by common knowledge, are geometrically planar surfaces. Take Fig.~\ref{fig::supp_nonplanar_gt_eval} as an example, the bottles are not considered as planes thus do not participate the plane fitting and annotation pipeline. However, the small faces of the bottle are essentially planes from a geometric perspective. If an algorithm detects and reconstructs these planes as in Fig.~\ref{fig::supp_nonplanar_gt_eval}, it should not get penalized in the evaluation protocol. 
\par
To this end, we propose a more robust evaluation pipeline to tolerate these cases. For every predicted plane mask which mostly locates at non-planar area~(over a threshold $50\%$), we first use the corresponding groundtruth depth to lift it to 3D, then apply least-square algorithm to fit its plane parameter. Then we calculate the average gap between reconstructed planar depth with its groundtruth. If the gap is below a certain threshold~($0.2m$), it will be regarded as a geometrically correct plane, and will not be penalized during evaluation. In this way, plane reconstruction quality will be assessed in a geometrically-preferred manner, regardless of its semantic category.
\par
To verify the difference between our proposed protocol and the original one, we display the comparisons on sparse-view reconstruction of two different methods on ScanNet~\cite{dai2017scannet} in Tab.~\ref{tab::twoview-AP-nonplanar-eval}. The numerical results of both methods get improved by tolerating the geometrically accurate planes without groundtruth annotations, which are treated as false positives in the original evaluation pipeline.

\subsection{More Experimental Results}
\paragraph{Sparse-view camera pose evaluation}
In Tab.~\ref{tab::twoview-camera}, we report the aligned pose evaluation following previous methods~\cite{agarwala2022planeformers, tan2022nope}. These data-driven methods~\cite{agarwala2022planeformers, tan2022nope} regress pose with learned plane embeddings trained on certain dataset, hindering the transferring capacity to novel scenes. We employ a robust SfM pipeline COLMAP~\cite{schonberger2016structure} with LoFTR~\cite{sun2021loftr} for keypoint detection and matching, showing superior generalizability across domains.
\begin{table}[]
\centering
\resizebox{1.0\linewidth}{!}{
\begin{tabular}{lcccc}
\toprule
\centering
\multirow{2}{*}{Settings / Metrics} & \multicolumn{4}{c}{Offset $\leq$ 1m, Normal $\leq$ 30°} \\ \cmidrule(r){2-5} 
                                  & All   & -Offset  & -Normal  & -Offset-Normal   \\

\midrule 
\multicolumn{5}{c}{NopeSAC~\cite{tan2022nope}} \\
\midrule
Origin & 2.47 & 3.17 & 5.13  & 7.09 \\
New & \textbf{2.55} & \textbf{3.27} & \textbf{5.34}  & \textbf{7.38} \\

\midrule 
\multicolumn{5}{c}{Ours} \\
\midrule
Origin & 12.73 & 15.67 & 15.35  & 19.47 \\
New & \textbf{15.30} & \textbf{18.82} & \textbf{18.48}  & \textbf{23.39} \\

\bottomrule
\end{tabular}
}

\caption{Comparison between two-view plane reconstruction on different evaluation protocol on ScanNet dataset.}
\label{tab::twoview-AP-nonplanar-eval}
\end{table}

\paragraph{Metric depth evaluation.}
To verify our plane reconstruction quality under metric depth, we evaluate the plane recall~\wrt the same depth discrepancy without scale alignment using our model with ZeroDepth~\cite{guizilini2023towards} as monocular input in Tab.~\ref{tab::metric_depth}. We can see our method achieves comparable performance on NYU~\cite{silberman2012indoor} with the learning-based counterparts, clearly better performance on Matterport3D~\cite{chang2017matterport3d}. On the outdoor dataset Synthia~\cite{ros2016synthia}, the other methods nearly fail on obtaining plausible plane geometry while our approach achieves feasible plane reconstruction even on the metric depth evaluation protocol. This demonstrates the promising potential of our paradigm to get desirable plane reconstruction with correct metric scale. One can believe that, with more powerful monocular depth estimation methods being explored and proposed, our method will benefit from the enhanced monocular input and achieve further favorable metric plane reconstruction.

\paragraph{Baselines with CRFs.} 
To verify the benefit brought by CRFs and ensure fair comparison by applying the same post-processing steps, we employ the same dense CRF on top of the output for each baseline counterparts. As shown in Tab.~\ref{tab::supp_crf}, applying CRFs only produces marginal effect on these methods. This is rationale since these learning-based methods have exploited sufficient image features by their powerful feature extraction backbone. Contrastively, our method benefits more from employing CRFs since our method is mainly geometry-driven by adopting monocular cues as input. CRFs can provide additional colour feature and assists our approach via better making use of the original image hints.

\paragraph{Ablation on distance threshold.} RANSAC-based methods are known to be sensitive to the hyper-parameters, especially the distance error threshold $\epsilon$. Our designs facilitate multiple point similarities within a graph to solve for the inlier labeling, thus to some extent robust to the threshold. We show the ablation of the distance error threshold $\epsilon$ in Tab.~\ref{tab::supp_ablation_threshold}.
\begin{table}[]
\centering
\resizebox{1.0\linewidth}{!}{
\begin{tabular}{lccccccccc}
\toprule
\centering
Metrics / Threshold & 0.001 & 0.002 & 0.005 & 0.01 & 0.02 & 0.05 & 0.1 & 0.2 & 0.5 \\
\midrule
SC~(↑) & 0.511 & 0.607 & 0.714 & 0.768 & 0.798 & \textbf{0.800} & 0.785 & 0.755 & 0.740 \\
RI~(↑) & 0.660 & 0.789 & 0.868 & 0.897 & \textbf{0.912} & 0.907 & 0.891 & 0.861 & 0.839 \\
VOI~(↓) & 1.788 & 1.535 & 1.204 & 1.023 & 0.910 & \textbf{0.878} & 0.909 & 0.988 & 1.029 \\
\bottomrule
\end{tabular}}
\caption{Ablation studies of RANSAC distance error threshold on ScanNet~\cite{dai2017scannet} dataset.}
\label{tab::supp_ablation_threshold}
\end{table}

\begin{figure*}[tb]
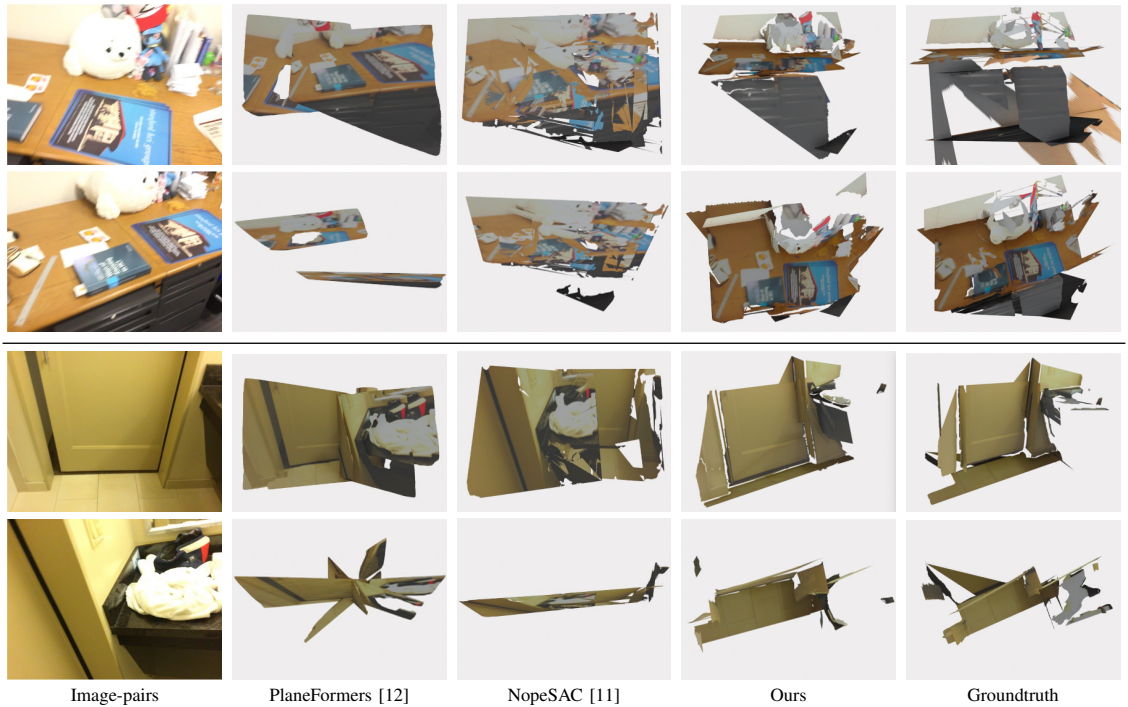

    \centering
    \scriptsize
    \setlength{\tabcolsep}{2pt}
    \begin{tabular}{ccccc}

    \centering
        {\includegraphics[width=0.16\linewidth]{figures/visualizations/supp_twoview/1-1.png}} &  
        {\includegraphics[width=0.16\linewidth]{figures/visualizations/supp_twoview/1-2.png}} &  
        {\includegraphics[width=0.16\linewidth]{figures/visualizations/supp_twoview/1-3.png}} & 
        {\includegraphics[width=0.16\linewidth]{figures/visualizations/supp_twoview/1-4.png}} &
        {\includegraphics[width=0.16\linewidth]{figures/visualizations/supp_twoview/1-5.png}}\\

        {\includegraphics[width=0.16\linewidth]{figures/visualizations/supp_twoview/2-1.png}} &  
        {\includegraphics[width=0.16\linewidth]{figures/visualizations/supp_twoview/2-2.png}} &  
        {\includegraphics[width=0.16\linewidth]{figures/visualizations/supp_twoview/2-3.png}} & 
        {\includegraphics[width=0.16\linewidth]{figures/visualizations/supp_twoview/2-4.png}} &
        {\includegraphics[width=0.16\linewidth]{figures/visualizations/supp_twoview/2-5.png}}\\
        \midrule

        {\includegraphics[width=0.16\linewidth]{figures/visualizations/supp_twoview/3-1.png}} &  
        {\includegraphics[width=0.16\linewidth]{figures/visualizations/supp_twoview/3-2.png}} &  
        {\includegraphics[width=0.16\linewidth]{figures/visualizations/supp_twoview/3-3.png}} & 
        {\includegraphics[width=0.16\linewidth]{figures/visualizations/supp_twoview/3-4.png}} &
        {\includegraphics[width=0.16\linewidth]{figures/visualizations/supp_twoview/3-5.png}}\\

        {\includegraphics[width=0.16\linewidth]{figures/visualizations/supp_twoview/4-1.png}} &  
        {\includegraphics[width=0.16\linewidth]{figures/visualizations/supp_twoview/4-2.png}} &  
        {\includegraphics[width=0.16\linewidth]{figures/visualizations/supp_twoview/4-3.png}} & 
        {\includegraphics[width=0.16\linewidth]{figures/visualizations/supp_twoview/4-4.png}} &
        {\includegraphics[width=0.16\linewidth]{figures/visualizations/supp_twoview/4-5.png}}\\

        Image-pairs &  PlaneFormers~\cite{agarwala2022planeformers} & NopeSAC~\cite{tan2022nope} & Ours & Groundtruth \\

    \end{tabular}
    \caption{More visualization of reconstructed mesh on two-view reconstruction on ScanNet~\cite{dai2017scannet}.}
    \label{fig::supp_twoview-scannet}
\end{figure*}
\begin{figure*}[th]
    \centering
    \scriptsize
    \setlength{\tabcolsep}{2pt}
    \begin{tabular}{cccc}

    \centering
        {\includegraphics[width=0.20\linewidth]{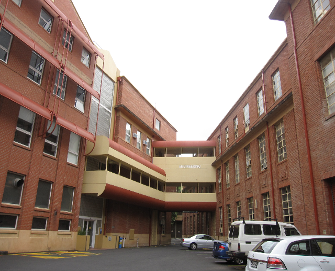}} &  
        {\includegraphics[width=0.20\linewidth]{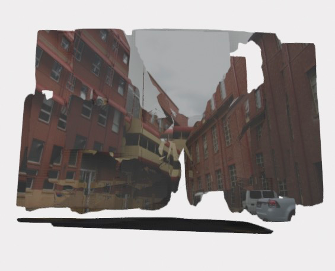}} &  
        {\includegraphics[width=0.20\linewidth]{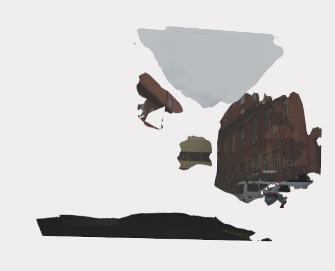}} & 
        {\includegraphics[width=0.20\linewidth]{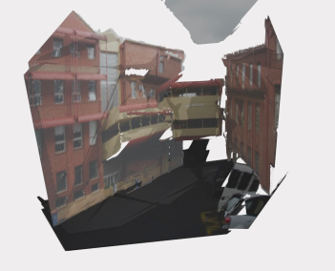}} \\

        {\includegraphics[width=0.20\linewidth]{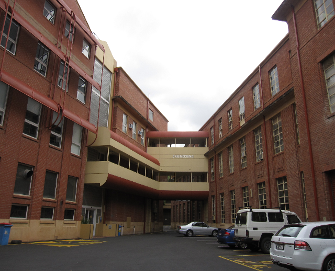}} &  
        {\includegraphics[width=0.20\linewidth]{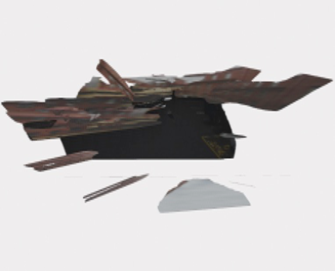}} &  
        {\includegraphics[width=0.20\linewidth]{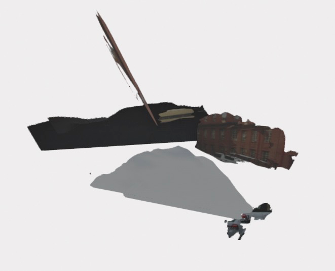}} & 
        {\includegraphics[width=0.20\linewidth]{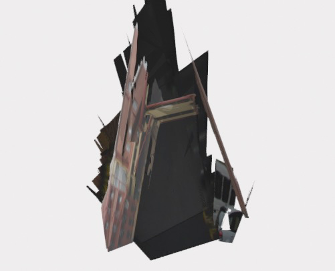}} \\
        \midrule

        {\includegraphics[width=0.20\linewidth]{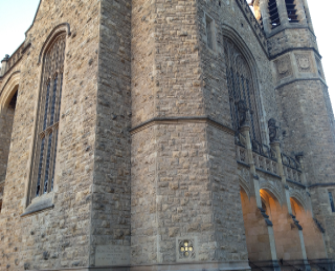}} &  
        {\includegraphics[width=0.20\linewidth]{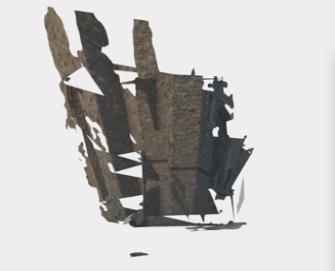}} &  
        {\includegraphics[width=0.20\linewidth]{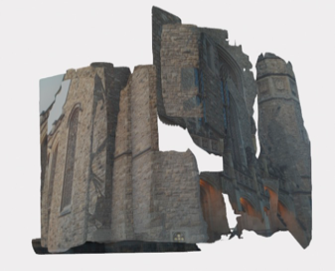}} & 
        {\includegraphics[width=0.20\linewidth]{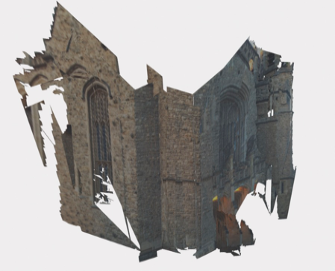}} \\

        {\includegraphics[width=0.20\linewidth]{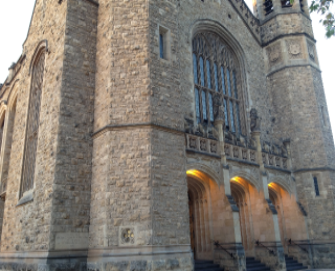}} &  
        {\includegraphics[width=0.20\linewidth]{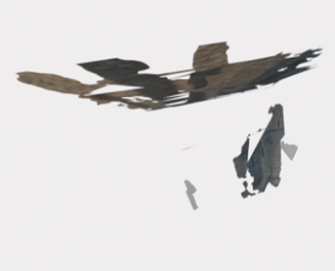}} &  
        {\includegraphics[width=0.20\linewidth]{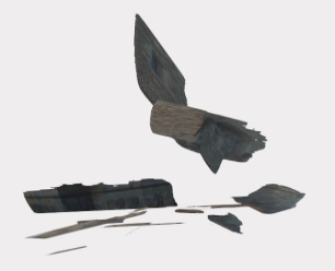}} & 
        {\includegraphics[width=0.20\linewidth]{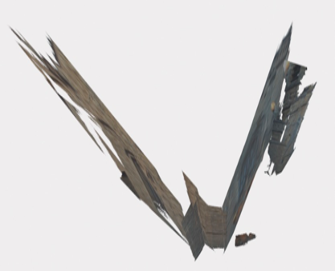}} \\

        Image-pairs &  PlaneFormers~\cite{agarwala2022planeformers} & NopeSAC~\cite{tan2022nope} & Ours \\

    \end{tabular}
    \caption{More visualization of reconstructed mesh on two-view reconstruction on wild~(Adelaide~\cite{wong2011dynamic}) data.}
    \label{fig::supp_twoview_adelaide}
\end{figure*}
\subsection{More Qualitative Results}

We show more qualitative comparison among different approaches in Fig.~\ref{fig::supp_singleview} for single-view testing datasets, Fig.~\ref{fig::supp_singleviewwild_3dpw_davis},~\ref{fig::supp_singleviewwild_sintel_kitti} for single-view wild datasets, Fig.~\ref{fig::supp_twoview-scannet} for two-view reconstruction on ScanNet dataset, and Fig.~\ref{fig::supp_twoview_adelaide} for two-view reconstruction on wild~(Adelaide) dataset.

\bibliographystyle{IEEEtran}
\bibliography{IEEEfull}

\begin{thebibliography}{10}
\providecommand{\url}[1]{#1}
\csname url@rmstyle\endcsname
\providecommand{\newblock}{\relax}
\providecommand{\bibinfo}[2]{#2}
\providecommand\BIBentrySTDinterwordspacing{\spaceskip=0pt\relax}
\providecommand\BIBentryALTinterwordstretchfactor{4}
\providecommand\BIBentryALTinterwordspacing{\spaceskip=\fontdimen2\font plus
\BIBentryALTinterwordstretchfactor\fontdimen3\font minus \fontdimen4\font\relax}
\providecommand\BIBforeignlanguage[2]{{%
\expandafter\ifx\csname l@#1\endcsname\relax
\typeout{** WARNING: IEEEtran.bst: No hyphenation pattern has been}%
\typeout{** loaded for the language `#1'. Using the pattern for}%
\typeout{** the default language instead.}%
\else
\language=\csname l@#1\endcsname
\fi
#2}}

\bibitem{sinha2009piecewise}
S.~Sinha, D.~Steedly, and R.~Szeliski, ``Piecewise planar stereo for image-based rendering,'' in \emph{ICCV}, 2009, pp. 1881--1888.

\bibitem{salas2014dense}
R.~F. Salas-Moreno, B.~Glocken, P.~H. Kelly, and A.~J. Davison, ``Dense planar slam,'' in \emph{ISMAR}.\hskip 1em plus 0.5em minus 0.4em\relax IEEE, 2014, pp. 157--164.

\bibitem{raposo2013plane}
C.~Raposo, M.~Louren{\c{c}}o, M.~Antunes, and J.~P. Barreto, ``Plane-based odometry using an rgb-d camera.'' in \emph{BMVC}, vol.~2, no.~5, 2013, p.~6.

\bibitem{furukawa2009manhattan}
Y.~Furukawa, B.~Curless, S.~M. Seitz, and R.~Szeliski, ``Manhattan-world stereo,'' in \emph{CVPR}, 2009, pp. 1422--1429.

\bibitem{fischler1981random}
M.~A. Fischler and R.~C. Bolles, ``Random sample consensus: a paradigm for model fitting with applications to image analysis and automated cartography,'' \emph{Communications of the ACM}, vol.~24, no.~6, pp. 381--395, 1981.

\bibitem{boykov1998markov}
Y.~Boykov, O.~Veksler, and R.~Zabih, ``Markov random fields with efficient approximations,'' in \emph{CVPR}, 1998, pp. 648--655.

\bibitem{liu2019planercnn}
C.~Liu, K.~Kim, J.~Gu, Y.~Furukawa, and J.~Kautz, ``Planercnn: 3d plane detection and reconstruction from a single image,'' in \emph{CVPR}, 2019, pp. 4450--4459.

\bibitem{qian2020learning}
Y.~Qian and Y.~Furukawa, ``Learning pairwise inter-plane relations for piecewise planar reconstruction,'' in \emph{ECCV}, 2020, pp. 330--345.

\bibitem{tan2021planetr}
B.~Tan, N.~Xue, S.~Bai, T.~Wu, and G.-S. Xia, ``Planetr: Structure-guided transformers for 3d plane recovery,'' in \emph{ICCV}, 2021, pp. 4186--4195.

\bibitem{liu2022planemvs}
J.~Liu, P.~Ji, N.~Bansal, C.~Cai, Q.~Yan, X.~Huang, and Y.~Xu, ``Planemvs: 3d plane reconstruction from multi-view stereo,'' in \emph{CVPR}, 2022, pp. 8665--8675.

\bibitem{tan2022nope}
B.~Tan, N.~Xue, T.~Wu, and G.-S. Xia, ``Nope-sac: Neural one-plane ransac for sparse-view planar 3d reconstruction,'' \emph{arXiv preprint arXiv:2211.16799}, 2022.

\bibitem{agarwala2022planeformers}
S.~Agarwala, L.~Jin, C.~Rockwell, and D.~F. Fouhey, ``Planeformers: From sparse view planes to 3d reconstruction,'' in \emph{ECCV}, 2022, pp. 192--209.

\bibitem{xie2022planarrecon}
Y.~Xie, M.~Gadelha, F.~Yang, X.~Zhou, and H.~Jiang, ``Planarrecon: Real-time 3d plane detection and reconstruction from posed monocular videos,'' in \emph{CVPR}, 2022, pp. 6219--6228.

\bibitem{dai2017scannet}
A.~Dai, A.~X. Chang, M.~Savva, M.~Halber, T.~Funkhouser, and M.~Nie{\ss}ner, ``Scannet: Richly-annotated 3d reconstructions of indoor scenes,'' in \emph{CVPR}, 2017, pp. 5828--5839.

\bibitem{chang2017matterport3d}
A.~Chang, A.~Dai, T.~Funkhouser, M.~Halber, M.~Niessner, M.~Savva, S.~Song, A.~Zeng, and Y.~Zhang, ``Matterport3d: Learning from rgb-d data in indoor environments,'' \emph{arXiv preprint arXiv:1709.06158}, 2017.

\bibitem{shi2023planerectr}
J.~Shi, S.~Zhi, and K.~Xu, ``Planerectr: Unified query learning for 3d plane recovery from a single view,'' in \emph{ICCV}, 2023, pp. 9377--9386.

\bibitem{jin2021planar}
L.~Jin, S.~Qian, A.~Owens, and D.~F. Fouhey, ``Planar surface reconstruction from sparse views,'' in \emph{ICCV}, 2021, pp. 12\,991--13\,000.

\bibitem{ranftl2020towards}
R.~Ranftl, K.~Lasinger, D.~Hafner, K.~Schindler, and V.~Koltun, ``Towards robust monocular depth estimation: Mixing datasets for zero-shot cross-dataset transfer,'' \emph{IEEE Transactions on Pattern Analysis and Machine Intelligence}, vol.~44, no.~3, pp. 1623--1637, 2020.

\bibitem{eftekhar2021omnidata}
A.~Eftekhar, A.~Sax, J.~Malik, and A.~Zamir, ``Omnidata: A scalable pipeline for making multi-task mid-level vision datasets from 3d scans,'' in \emph{ICCV}, 2021, pp. 10\,786--10\,796.

\bibitem{bhat2023zoedepth}
S.~F. Bhat, R.~Birkl, D.~Wofk, P.~Wonka, and M.~M{\"u}ller, ``Zoedepth: Zero-shot transfer by combining relative and metric depth,'' \emph{arXiv preprint arXiv:2302.12288}, 2023.

\bibitem{Yu2022MonoSDF}
Z.~Yu, S.~Peng, M.~Niemeyer, T.~Sattler, and A.~Geiger, ``Monosdf: Exploring monocular geometric cues for neural implicit surface reconstruction,'' \emph{NeurIPS}, 2022.

\bibitem{zhu2023nicer}
Z.~Zhu, S.~Peng, V.~Larsson, Z.~Cui, M.~R. Oswald, A.~Geiger, and M.~Pollefeys, ``Nicer-slam: Neural implicit scene encoding for rgb slam,'' \emph{arXiv preprint arXiv:2302.03594}, 2023.

\bibitem{ros2016synthia}
G.~Ros, L.~Sellart, J.~Materzynska, D.~Vazquez, and A.~M. Lopez, ``The synthia dataset: A large collection of synthetic images for semantic segmentation of urban scenes,'' in \emph{CVPR}, 2016, pp. 3234--3243.

\bibitem{perazzi2016benchmark}
F.~Perazzi, J.~Pont-Tuset, B.~McWilliams, L.~Van~Gool, M.~Gross, and A.~Sorkine-Hornung, ``A benchmark dataset and evaluation methodology for video object segmentation,'' in \emph{CVPR}, 2016, pp. 724--732.

\bibitem{barath2018graph}
D.~Barath and J.~Matas, ``Graph-cut ransac,'' in \emph{CVPR}, 2018, pp. 6733--6741.

\bibitem{krahenbuhl2011efficient}
P.~Kr{\"a}henb{\"u}hl and V.~Koltun, ``Efficient inference in fully connected crfs with gaussian edge potentials,'' \emph{NeurIPS}, vol.~24, 2011.

\bibitem{gallup2010piecewise}
D.~Gallup, J.-M. Frahm, and M.~Pollefeys, ``Piecewise planar and non-planar stereo for urban scene reconstruction,'' in \emph{CVPR}, 2010, pp. 1418--1425.

\bibitem{liu2018planenet}
C.~Liu, J.~Yang, D.~Ceylan, E.~Yumer, and Y.~Furukawa, ``Planenet: Piece-wise planar reconstruction from a single rgb image,'' in \emph{CVPR}, 2018, pp. 2579--2588.

\bibitem{yang2018recovering}
F.~Yang and Z.~Zhou, ``Recovering 3d planes from a single image via convolutional neural networks,'' in \emph{ECCV}, 2018, pp. 85--100.

\bibitem{yu2019single}
Z.~Yu, J.~Zheng, D.~Lian, Z.~Zhou, and S.~Gao, ``Single-image piece-wise planar 3d reconstruction via associative embedding,'' in \emph{CVPR}, 2019, pp. 1029--1037.

\bibitem{ye2023self}
B.~Ye, S.~Liu, X.~Li, and M.-H. Yang, ``Self-supervised super-plane for neural 3d reconstruction,'' in \emph{CVPR}, 2023, pp. 21\,415--21\,424.

\bibitem{eigen2014depth}
D.~Eigen, C.~Puhrsch, and R.~Fergus, ``Depth map prediction from a single image using a multi-scale deep network,'' \emph{NeurIPS}, vol.~27, 2014.

\bibitem{eigen2015predicting}
D.~Eigen and R.~Fergus, ``Predicting depth, surface normals and semantic labels with a common multi-scale convolutional architecture,'' in \emph{ICCV}, 2015, pp. 2650--2658.

\bibitem{bhat2021adabins}
S.~F. Bhat, I.~Alhashim, and P.~Wonka, ``Adabins: Depth estimation using adaptive bins,'' in \emph{CVPR}, 2021, pp. 4009--4018.

\bibitem{li2023depthformer}
Z.~Li, Z.~Chen, X.~Liu, and J.~Jiang, ``Depthformer: Exploiting long-range correlation and local information for accurate monocular depth estimation,'' \emph{Machine Intelligence Research}, pp. 1--18, 2023.

\bibitem{agarwal2023attention}
A.~Agarwal and C.~Arora, ``Attention attention everywhere: Monocular depth prediction with skip attention,'' in \emph{WACV}, 2023, pp. 5861--5870.

\bibitem{bansal2016marr}
A.~Bansal, B.~Russell, and A.~Gupta, ``Marr revisited: 2d-3d alignment via surface normal prediction,'' in \emph{CVPR}, 2016, pp. 5965--5974.

\bibitem{geiger2013vision}
A.~Geiger, P.~Lenz, C.~Stiller, and R.~Urtasun, ``Vision meets robotics: The kitti dataset,'' \emph{The International Journal of Robotics Research}, vol.~32, no.~11, pp. 1231--1237, 2013.

\bibitem{silberman2012indoor}
N.~Silberman, D.~Hoiem, P.~Kohli, and R.~Fergus, ``Indoor segmentation and support inference from rgbd images.'' \emph{ECCV}, vol. 7576, pp. 746--760, 2012.

\bibitem{song2015sun}
S.~Song, S.~P. Lichtenberg, and J.~Xiao, ``Sun rgb-d: A rgb-d scene understanding benchmark suite,'' in \emph{CVPR}, 2015, pp. 567--576.

\bibitem{ranftl2021vision}
R.~Ranftl, A.~Bochkovskiy, and V.~Koltun, ``Vision transformers for dense prediction,'' in \emph{ICCV}, 2021, pp. 12\,179--12\,188.

\bibitem{dosovitskiy2020image}
A.~Dosovitskiy, L.~Beyer, A.~Kolesnikov, D.~Weissenborn, X.~Zhai, T.~Unterthiner, M.~Dehghani, M.~Minderer, G.~Heigold, S.~Gelly, \emph{et~al.}, ``An image is worth 16x16 words: Transformers for image recognition at scale,'' \emph{arXiv preprint arXiv:2010.11929}, 2020.

\bibitem{guizilini2023towards}
V.~Guizilini, I.~Vasiljevic, D.~Chen, R.~Ambruș, and A.~Gaidon, ``Towards zero-shot scale-aware monocular depth estimation,'' in \emph{ICCV}, 2023, pp. 9233--9243.

\bibitem{yin2023metric3d}
W.~Yin, C.~Zhang, H.~Chen, Z.~Cai, G.~Yu, K.~Wang, X.~Chen, and C.~Shen, ``Metric3d: Towards zero-shot metric 3d prediction from a single image,'' in \emph{ICCV}, 2023, pp. 9043--9053.

\bibitem{torr2002bayesian}
P.~Torr, ``Bayesian model estimation and selection for epipolar geometry and generic manifold fitting,'' \emph{International Journal of Computer Vision}, vol.~50, no.~1, p.~35, 2002.

\bibitem{torr2000mlesac}
P.~H. Torr and A.~Zisserman, ``Mlesac: A new robust estimator with application to estimating image geometry,'' \emph{Computer Vision and Image Understanding}, vol.~78, no.~1, pp. 138--156, 2000.

\bibitem{barath2020magsac++}
D.~Barath, J.~Noskova, M.~Ivashechkin, and J.~Matas, ``Magsac++, a fast, reliable and accurate robust estimator,'' in \emph{CVPR}, 2020, pp. 1304--1312.

\bibitem{tomasi1998bilateral}
C.~Tomasi and R.~Manduchi, ``Bilateral filtering for gray and color images,'' in \emph{ICCV}.\hskip 1em plus 0.5em minus 0.4em\relax IEEE, 1998, pp. 839--846.

\bibitem{kolmogorov2004energy}
V.~Kolmogorov and R.~Zabin, ``What energy functions can be minimized via graph cuts?'' \emph{IEEE Transactions on Pattern Analysis and Machine Intelligence}, vol.~26, no.~2, pp. 147--159, 2004.

\bibitem{schonberger2016structure}
J.~L. Schonberger and J.-M. Frahm, ``Structure-from-motion revisited,'' in \emph{CVPR}, 2016, pp. 4104--4113.

\bibitem{sun2021loftr}
J.~Sun, Z.~Shen, Y.~Wang, H.~Bao, and X.~Zhou, ``Loftr: Detector-free local feature matching with transformers,'' in \emph{CVPR}, 2021, pp. 8922--8931.

\bibitem{sturm2012benchmark}
J.~Sturm, N.~Engelhard, F.~Endres, W.~Burgard, and D.~Cremers, ``A benchmark for the evaluation of rgb-d slam systems,'' in \emph{IROS}.\hskip 1em plus 0.5em minus 0.4em\relax IEEE, 2012, pp. 573--580.

\bibitem{vonMarcard2018}
T.~von Marcard, R.~Henschel, M.~Black, B.~Rosenhahn, and G.~Pons-Moll, ``Recovering accurate 3d human pose in the wild using imus and a moving camera,'' in \emph{ECCV}, sep 2018.

\bibitem{kirillov2023segment}
A.~Kirillov, E.~Mintun, N.~Ravi, H.~Mao, C.~Rolland, L.~Gustafson, T.~Xiao, S.~Whitehead, A.~C. Berg, W.-Y. Lo, \emph{et~al.}, ``Segment anything,'' \emph{arXiv preprint arXiv:2304.02643}, 2023.

\bibitem{wong2011dynamic}
H.~S. Wong, T.-J. Chin, J.~Yu, and D.~Suter, ``Dynamic and hierarchical multi-structure geometric model fitting,'' in \emph{ICCV}.\hskip 1em plus 0.5em minus 0.4em\relax IEEE, 2011, pp. 1044--1051.

\end{thebibliography}

\end{document}